\documentclass[letterpaper]{article} %
\usepackage[preprint]{aaai2027}  %
\usepackage[hyphens]{url}  %
\usepackage{graphicx} %
\usepackage{natbib}  %
\usepackage{caption} %
\usepackage{algorithm}
\usepackage{algorithmic}
\usepackage{booktabs}
\usepackage{amsmath,amssymb}

\newcommand{\omcr}{OMCR}

\newcommand{\scg}{Subspace Candidate Gate}
\newcommand{\scgabbr}{SCG}
\newcommand{\retreadout}{SCG-Ret}
\newcommand{\reconreadout}{SCG-Recon}
\newcommand{\fulldiag}{FullDiag}
\newcommand{\things}{\textsc{THINGS-EEG}}

\title{SPSC: Zero-Shot Cross-Subject EEG-to-Image Reconstruction via Semantic Preservation and Subspace Calibration}
\author{Yutang Lu, Shuai Jiang, Fei Su, Zhicheng Zhao}
\affiliations{Beijing University of Posts and Telecommunications}

\begin{document}

\maketitle

\begin{abstract}
Pretrained vision-language models and diffusion generators have driven rapid progress in brain-to-image generation, while EEG-based reconstruction is still constrained by low signal-to-noise ratio and cross-subject semantic shifts.
Zero-shot cross-subject reconstruction poses even greater challenges: EEG semantic embeddings learned from source subjects deviate from the manifold of image embeddings of the unseen target subject, particularly when neither target labels nor target image embeddings are available.
To address this issue, we introduce the Semantic Preservation and Subspace Calibration (SPSC) framework, which first uses Object Mask Contrastive Regularization (\omcr{}) with object identity for backend training, and then constructs \scg{} (\scgabbr{}) to achieve label-free semantic correction for target subjects during inference.
Specifically, \omcr{} regularizes the conditioning space with object-level positive samples to preserve object-level feature neighborhoods within the reconstruction prior and low-level branch, while \scgabbr{} exploits training image prototypes and batch statistics of unlabeled target EEG to select candidates for each task within subspaces of image prototypes.
Under the leave-one-subject-out (LOSO) setting across ten subjects on the \things{} dataset, SPSC improves the cross-subject reconstruction from 0.552 to 0.713 on CLIP and from 0.612 to 0.745 on Alex.5.
These significant gains indicate that the semantically calibrated conditions can better support generation for the target subjects under label-free inference.
\end{abstract}

\section{Introduction}

Visual neural decoding aims to recover perceived image content from brain activity.
Early work established the feasibility of image reconstruction from fMRI because its spatial specificity provides stable visual responses
\citep{naselaris2009bayesian}.
Recent systems further connect neural activity with pretrained vision-language representations and diffusion generators, enabling reconstruction of natural images from fMRI and MEG
\citep{takagi2023high,ozcelik2023natural,chen2023mindvis,scotti2023mindeye,benchetrit2024brain}.
However, these results show that a powerful generator is not sufficient by itself: the neural signal must be converted into semantic conditions that remain meaningful to the generator.

EEG is well-suited for this task because it is portable, inexpensive, and temporally precise.
Large-scale datasets such as \things{} provide dense responses to diverse natural images and make EEG-based visual decoding measurable at the object concept level
\citep{hebart2019things,grootswagers2022human,gifford2022large}.
Recent EEG methods align EEG signals with pretrained visual embeddings and use the aligned representation for retrieval, generation, or both tasks
\citep{song2024nice,bai2023dreamdiffusion,singh2023eeg2image,li2024atm,wu2025ubp,zhang2025neurobridge,dai2026avde}.
These studies establish pretrained visual spaces as an effective interface for EEG decoding, but this interface remains vulnerable when applied to unseen subjects without labeled calibration images.

However, cross-subject reconstruction faces two core challenges.
First, anatomical, physiological, and recording differences induce subject-dependent EEG shifts, thus a source-trained projector may yield target embeddings whose global distribution and local coordinates shift away from the image embedding manifold
\cite{kobler2022eegdomain,huang2025need}. %
Second, existing EEG--image objectives mainly optimize one-to-one trial--image alignment
\cite{song2024nice,li2024atm,zhang2025neurobridge},
without explicitly preserving relationships among images sharing the same object concept.
Insights from representational geometry offer a feasible solution: categorical and exemplar-level structure can be recovered from single-trial EEG recordings
\cite{Kaneshiro2015single},
while fMRI studies reveal that visual cortical responses can be aligned into a shared representational space across different individuals, even given idiosyncratic response topographies
\cite{HAXBY2011HighDimensional}.
These findings motivate us to preserve object-level neighborhoods throughout training and calibrate subject embeddings within a shared visual semantic space at inference.

Therefore, we propose the Semantic Preservation and Subspace Calibration (SPSC) framework for zero-shot cross-subject EEG-to-image reconstruction, which introduces two core modules: Object Mask Contrastive Regularization (OMCR) and Subspace Candidate Gate (SCG).
\omcr{} regularizes this conditioning space with object-level positive samples, thus it maintains pairwise EEG-image alignment and ensures EEG responses corresponding to the same object concept remain semantically adjacent.
At inference time, \scgabbr{} calibrates the target subject's EEG semantic embeddings after the projector.
It builds candidate semantic conditions within PCA subspaces of image prototypes using training image prototypes and batch statistics of unlabeled target EEG data. A label-free confidence gate then selects a valid candidate for retrieval or reconstruction.

Empirically, SPSC improves cross-subject reconstruction performance under a leave-one-subject-out (LOSO) protocol with ten subjects on the \things{} dataset:
raising CLIP from 0.552 to 0.713 and Alex.5 from 0.612 to 0.745.
Detailed analyses on LOSO Sub-08 show that low-rank candidates in the PCA subspace of image prototypes yield smoother and more reliable generative conditions, while higher-rank candidates retain fine-grained discriminative structures for the retrieval task.

Our contributions are the following four aspects:
\begin{itemize}
\item We introduce a novel zero-shot cross-subject EEG-to-image reconstruction framework that significantly boosts reconstruction performance by calibrating semantic conditions for the target subject without target labels or target image embeddings.
\item We propose an Object Mask Contrastive Regularization (\omcr{}) that preserves semantic neighborhood relationships beyond simple pairwise EEG--image alignment via regularizing conditioning space with object-level positive samples.
\item We design the Subspace Candidate Gate (\scgabbr{}), a label-free candidate gate operating within image subspaces, which leverages training image prototypes and unlabeled target EEG batch statistics to mitigate the subject shift.
\item We conduct comprehensive evaluations on the \things{} dataset, including reconstruction and retrieval tasks. The results validate that the proposed framework outperforms the SOTA methods, and subspace analyses demonstrate that low-rank candidates in the PCA subspace of image prototypes deliver superior conditioning for reconstruction, whereas mid/high-rank candidates better preserve discriminative structures for retrieval.
\end{itemize}

\section{Related Work}

\paragraph{Brain-to-image reconstruction.}
Modern brain-to-image systems commonly use pretrained visual priors.
A neural encoder predicts visual representations, which are then decoded into images by diffusion or unCLIP style generators
\citep{takagi2023high,ozcelik2023natural,scotti2023mindeye,benchetrit2024brain,ramesh2022unclip,rombach2022ldm}.
This architecture decouples neural encoding and image synthesis, enabling the generator to inherit large-scale pretrained semantic and visual knowledge.
However, cross-subject EEG-to-image reconstruction remains largely underexplored, especially in scenarios without labeled calibration data from the unseen target subject.
To this end, we propose a zero-shot cross-subject EEG-to-image reconstruction method via semantic preservation and subspace calibration.

\paragraph{EEG visual decoding.}
Existing EEG-to-image reconstruction and retrieval methods such as NICE, DreamDiffusion, EEG2IMAGE, ATM, UBP, NeuroBridge, and AVDE align EEG responses with pretrained visual representations
\citep{song2024nice,bai2023dreamdiffusion,singh2023eeg2image,li2024atm,wu2025ubp,zhang2025neurobridge,dai2026avde}.
Although these methods enable EEG-based visual reconstruction, most training objectives emphasize paired alignment between single EEG trials and individual image embeddings.
However, for datasets containing multiple images per object concept, training solely on isolated pairs restricts the conditioning space to discrete matched pairs, failing to form coherent object-level semantic structures.
Accordingly, we introduce \omcr{} to address this issue by constructing intra-batch positive samples based on shared object identities.

\paragraph{Cross-subject adaptation.}
Cross-subject transfer has long been a fundamental challenge in neural decoding.
For fMRI and related modalities, subject alignment techniques reduce data requirements and improve cross-subject generalization
\citep{scotti2024mindeye2,wang2024mindbridge,gong2025mindtuner}.
However, for EEG, variations stemming from anatomical differences, electrode placement, and trial noise make cross-subject transfer more fragile; NEED also highlights cross-task generalization for EEG-based visual reconstruction \citep{huang2025need}.
The proposed \scgabbr{} further complements these efforts by operating calibration after the EEG projector: it adjusts EEG embeddings toward the manifold spanned by training image prototypes, relying solely on batch statistics extracted from unlabeled target EEG data.

\section{Method}

\begin{figure*}[t]
    \centering
    \includegraphics[width=\textwidth]{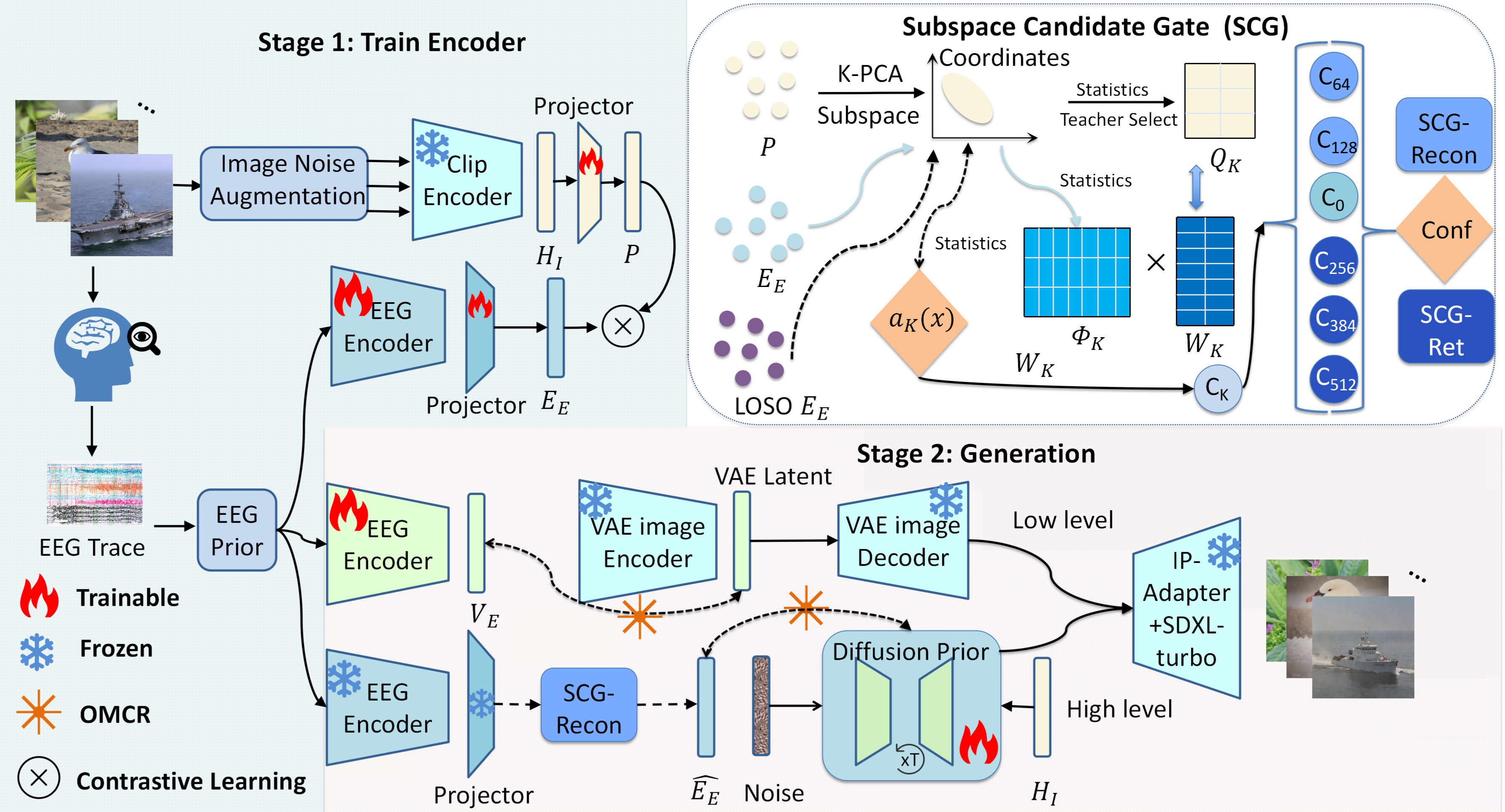}
\caption{SPSC framework for zero-shot cross-subject EEG-to-image reconstruction. In the stage 1, we train the EEG encoder with the linear projector via constrastive learning and transform  EEG to semantic vectors $E_E$ ($x_{i,s}$).
Subsequently, SCG uses training image prototypes $P$ and LOSO $E_E$ statistics extracted from the target subject to construct semantic candidates $C_K$ in PCA subspaces of image prototypes, then selects semantic vectors in SCG-Recon or SCG-Ret through a label-free confidence gate. In the stage 2, we adopt SCG-Recon output $\widehat{E_E}$ for image generation, while \omcr{} preserves object-level neighborhoods within CLIP embeddings $H_I$ of the diffusion prior and the low-level VAE latent features $V_E$.}
    \label{fig:framework}
\end{figure*}

Figure~\ref{fig:framework} organizes SPSC around two core problems of zero-shot cross-subject reconstruction.
Firstly, during generation backend training, \omcr{} preserves object-level neighborhoods so that the generator receives semantic conditions with stable object structure.
Secondly, during target inference, \scgabbr{} calibrates post-projector EEG semantic vectors from unseen subjects using only training image prototypes and unlabeled target batch statistics.
Since retrieval and reconstruction prioritize different semantic scales, \scgabbr{} exposes two readouts from the same calibrated candidate pool: \retreadout{} for discriminative retrieval and \reconreadout{} for stable semantic conditioning in reconstruction.
These two modules act at different stages of the pipeline: \omcr{} regularizes the reconstruction backend, whereas \scgabbr{} only edits the semantic condition fed into the diffusion prior and generator.

\subsection{Problem Setup}

The reconstruction pipeline first maps EEG to a semantic vector $x\in\mathbb{R}^d$, and then predicts the CLIP image embedding $\hat{h}$ and low-level latent $\hat{z}$ for image generation.
This design facilitates the disentanglement between high-level semantic conditioning and low-level image structure.

In cross-subject reconstruction, one subject is held out as the unseen target test subject.
During inference, only unlabeled batch statistics from the target subject's EEG semantic vectors are available, while SCG branch adaptation and hyperparameter tuning rely solely on source subjects.
Thus, this corresponds to an unlabeled test-batch adaptation setting: target EEG samples are observed in batches, while their stimulus identities and image embeddings are only used for evaluation after inference.

\subsection{Subspace Candidate Gate (SCG)}

Aiming to calibrate target-subject semantics after the EEG projector,
\scgabbr{} first learns the optimal transformation that aligns EEG semantic coordinates with the image prototype manifold using data from source subjects.
At target inference, \scgabbr{} instantiates the same correction paradigm using unlabeled target EEG batch statistics and selects one complete semantic candidate for each individual sample.
Note that SCG only edits semantic conditions output by the frozen projector;
\retreadout{} and \reconreadout{} serve as two task-aware readouts derived from the same candidate family: \retreadout{} exposes mid/high-rank candidates for retrieval, whereas \reconreadout{} leverages low-rank candidates for reconstruction.

Let $N_{\mathrm{tr}}$ be the total number of training stimuli.
Let $x_{i,s}\in\mathbb{R}^{D}$ be the EEG semantic vector output by the projector for the $i$-th stimulus from the $s$-th subject, where $D=1024$.
Each stimulus is associated with an image prototype $p_i$, which refers to the projected semantic embedding of its corresponding training image.
\begin{equation}
P=\{p_i\}_{i=1}^{N_{\mathrm{tr}}}.
\end{equation}
The prototype bank $P$ provides the image reference used by \scgabbr{} to compute PCA bases, distribution statistics, branch fitting, and candidate scoring.

\paragraph{Shared subspace and Correlation Alignment (CORAL) variants.}
The shared subspace determines the image-side coordinates that \scgabbr{} is allowed to modify. Accordingly, the PCA basis $B_K\in\mathbb{R}^{D\times K}$ is fitted only on the training prototype bank $P$, where $K\in\{64,128,256,384,512\}$.
For any semantic vector $x$, we define its coordinate in the PCA subspace of image prototypes and corresponding decoding mapping as
\begin{equation}
u_K(x)=B_K^\top(x-\mu_P),\qquad
d_K(u)=\mu_P+B_Ku .
\end{equation}
The prototype coordinate batch $V^K=\{u_K(p_i)\}_{i=1}^{N_{\mathrm{tr}}}$ provides the image-side statistics used to calibrate EEG coordinates within the rank-$K$ subspace.
Given a reference coordinate batch $R$ and a target coordinate batch $V$, we denote $\mu_R,\Sigma_R$ and $\mu_V,\Sigma_V$ as their respective mean and covariance.
At each rank, the regularized CORAL operator aligns EEG coordinate statistics to image prototype statistics by whitening $z$ with the reference covariance and recoloring it with the target covariance \citep{Coral}:
\begin{align}
\operatorname{CORAL}_{r}(z;R,V)
&=(z-\mu_R)(\Sigma_R+rI)^{-\frac{1}{2}} \nonumber\\
&\quad\cdot(\Sigma_V+rI)^{\frac{1}{2}}+\mu_V .
\end{align}
SCG uses an interpolated CORAL variant
\begin{equation}
T_{r,\rho}(z;R,V)
=z+\rho\left(\operatorname{CORAL}_{r}(z;R,V)-z\right),
\end{equation}
where $r$ controls covariance regularization and $\rho$ constrains the strength of alignment toward statistics of image prototype coordinates.

\paragraph{Subspace branch fitting.}
To learn a reliable semantic correction rule, we exploit the paired EEG-image evidence from source subjects.
For this purpose, \scgabbr{} fits a ridge mapper for each rank across all source subjects to shift EEG coordinates toward image prototype coordinates.
Then, to quantify how source EEG coordinates move toward the prototype coordinate batch $V^K$, we utilize the EEG coordinate batch $U_s^K=\{u_K(x_{i,s})\}_{i=1}^{N_{\mathrm{tr}}}$ from source subject $s$, which provides the subject statistics.
For each rank $K$, \scgabbr{} builds CORAL teacher variants under image prototype statistics:
\begin{equation}
t^{K,r,\rho}_{i,s}=T_{r,\rho}(u_K(x_{i,s});U_s^K,V^K).
\end{equation}
Among all variants, \scgabbr{} retains the variant whose decoded vector obtains the highest cosine similarity with the training prototype bank.
The ridge input $\phi_K$ contains a subject-normalized sample coordinate, the subject-specific mean and log standard deviation within the same subspace, along with a constant bias term.
This supplies the mapper both the per-sample coordinates and a concise representation of the subject-level distribution shift.
Stacking $\phi_K$ over all source samples yields the matrix $\Phi_K$, and stacking the selected teacher coordinates generates $Q_K$.
The parameters $W_K$ of this mapping branch are fitted by ridge regression:
\begin{equation}
W_K =
\arg\min_W
\left\|\Phi_K W-Q_K\right\|_F^2
\;+\;
\lambda\left\|W\right\|_F^2 .
\end{equation}
In implementation, we adopt the closed-form ridge solution with $\lambda=1$.
The fitted $W_K$ will project subject-normalized EEG subspace features into $K$-dimensional coordinates aligned with image prototypes.

\paragraph{Candidate construction for the target subject.}
Given an unlabeled EEG batch $X$, the pipeline for constructing target candidate construction is divided into three distinct components.
The first candidate $c_0$ corresponds to a full-dimensional diagonal alignment (\fulldiag{}), which eliminates global mismatch in mean and feature scale:
\begin{equation}
c_0(x)=\mu_P+\frac{\sigma_P}{\sigma_X+\epsilon}\odot(x-\mu_X),
\end{equation}
where $\mu_P,\sigma_P$ are prototype statistics and $\mu_X,\sigma_X$ are statistics of the target EEG test batch.
For each rank $K$, we define $U_X^K=\{u_K(x):x\in X\}$.
The CORAL anchor will shift the features of the target batch to a region close to the image prototype distribution within a selected subspace:
\begin{equation}
a_K^{r,\rho}(x)=d_K\!\left(T_{r,\rho}(u_K(x);U_X^K,V^K)\right).
\end{equation}
Here, the anchor variant achieving the highest cosine similarity with the training prototype bank is denoted as $a_K(x)$.
The target batch supplies the same mean and scale features used by the ridge mapper, producing an image-aligned coordinate prediction
$\hat{u}_K(x)=\phi_K(u_K(x))W_K$.
The final rank-$K$ candidate $c_K(x)$ applies a controlled residual $\delta_K(x)$ around the selected anchor $a_K(x)$, which are defined as follows:
\begin{align}
\delta_K(x) &= \hat{u}_K(x)-B_K^\top(a_K(x)-\mu_P),\\
c_K(x) &= a_K(x)+\beta B_K\delta_K(x).
\end{align}
The anchor handles distribution-level shift, while the residual carries the source-learned correction; the small blend $\beta$ preserves the target sample geometry within the selected semantic scale.

\paragraph{Label-free gate.}
At selection time, \scgabbr{} performs sample-wise routing over complete candidate vectors via prototype bank confidence:
\begin{equation}
C(x)=\{c_0(x),c_{64}(x),c_{128}(x),c_{256}(x),c_{384}(x),c_{512}(x)\}.
\end{equation}
Each $c_K(x)\in\mathbb{R}^{D}$ is a complete semantic condition produced by branch $K$.
Let $s^{K}_{(1)}\geq\cdots\geq s^{K}_{(10)}$ be the top 10 cosine similarities between $c_K(x)$ and prototypes in $P$.
The confidence score is
\begin{equation}
\operatorname{Conf}_K(x)=s^K_{(1)}+2\left(s^K_{(1)}-s^K_{(10)}\right),
\end{equation}
where, the nearest prototype term $s^K_{(1)}$ measures manifold proximity, and the 2nd term tail gap measures local sharpness.
The selector exports the eligible candidate $\hat{c}(x)$ with the highest confidence:
\begin{equation}
K^*(x)=\arg\max_{K\in A}\operatorname{Conf}_K(x),
\hat{c}(x)=c_{K^*(x)}(x).
\end{equation}
Subsequently, we set $A=\{0,256,384,512\}$ to define \retreadout{}, and set $A=\{0,64,128\}$ to define \reconreadout{}.
Both readouts are adopted to evaluate the same candidate pool: the retrieval readout retains higher-rank candidates for fine semantic discrimination, whereas the reconstruction readout restricts selection to lower-rank candidates that provide smoother diffusion conditioning.

\begin{table*}[t]
\centering
\small
\setlength{\tabcolsep}{1mm}
\begin{tabular*}{\textwidth}{@{\extracolsep{\fill}}ll|ccccccccccc@{}}
\toprule
Method & Metric (\%) & S01 & S02 & S03 & S04 & S05 & S06 & S07 & S08 & S09 & S10 & Mean \\
\midrule
NICE & Top-1 & 7.60 & 5.90 & 6.00 & 6.30 & 4.40 & 5.60 & 5.60 & 6.30 & 5.70 & 8.40 & 6.20 \\
{\cite{song2024nice}} & Top-5 & 22.8 & 20.5 & 22.3 & 20.7 & 18.3 & 22.2 & 19.7 & 22.0 & 17.6 & 28.3 & 21.4 \\
\midrule
ATM & Top-1 & 10.5 & 7.10 & 11.9 & 14.7 & 7.00 & 11.1 & \underline{16.1} & 15.0 & 4.90 & 20.5 & 11.9 \\
{\cite{li2024atm}} & Top-5 & 26.8 & 24.8 & 33.8 & 39.4 & 23.9 & 35.8 & \underline{43.5} & \underline{40.3} & 22.7 & 46.5 & 33.8 \\
\midrule
AVDE & Top-1 & \underline{14.1} & 17.0 & 9.10 & 15.2 & 12.5 & \underline{17.3} & 7.40 & \underline{18.5} & \underline{13.2} & 18.0 & 14.3 \\
{\cite{dai2026avde}} & Top-5 & 32.2 & 38.4 & 21.8 & 32.5 & 32.4 & \underline{38.6} & 20.4 & 40.1 & \underline{33.6} & 39.3 & 32.9 \\
\midrule
NeuroBridge (ViT-H-14) & Top-1 & 14.0 & \underline{25.5} & \underline{14.5} & \underline{15.5} & \underline{14.5} & 13.5 & 15.5 & 15.5 & 12.0 & \underline{23.5} & \underline{16.4} \\
{\cite{zhang2025neurobridge}} & Top-5 & \underline{39.0} & \underline{50.5} & \underline{33.0} & \underline{39.5} & \underline{36.5} & 37.0 & 40.0 & 36.0 & 30.5 & \underline{52.5} & \underline{39.5} \\
\midrule
\retreadout{} (Ours) & Top-1 & \textbf{24.5} & \textbf{28.0} & \textbf{20.5} & \textbf{18.0} & \textbf{19.0} & \textbf{27.0} & \textbf{22.0} & \textbf{24.0} & \textbf{23.0} & \textbf{30.5} & \textbf{23.7} \\
 & Top-5 & \textbf{47.5} & \textbf{53.5} & \textbf{47.5} & \textbf{42.0} & \textbf{44.0} & \textbf{54.5} & \textbf{46.5} & \textbf{46.0} & \textbf{48.0} & \textbf{62.5} & \textbf{49.2} \\
\bottomrule
\end{tabular*}
\caption{Cross-subject 200-way retrieval accuracy under ten-subject LOSO. \textbf{Best} and \underline{second best} are highlighted.}
\label{tab:retrieval-full}
\end{table*}

\subsection{Object Mask Contrastive Regularization (OMCR)}

\omcr{} focuses on addressing the existing training-side limitation of pairwise EEG--image alignment: different images from the same object concept can be treated as unrelated negatives.
Given predicted features $\hat{r}_i$, target features $r_i$, and object labels $y_i$ in a batch $B$, it uses object labels to preserve object-level neighborhoods while retaining the paired image target.
The paired cosine anchor loss $\mathcal{L}_{\mathrm{cos}}$ is defined as:
\begin{equation}
\mathcal{L}_{\mathrm{cos}}(\hat{r}, r)
= \frac{1}{B}\sum_{i=1}^B \left(1-\cos(\hat{r}_i,r_i)\right).
\end{equation}
Object labels build the mask $M_{ij}=\mathbf{1}[y_i=y_j]$ in training batches.
The final loss $\mathcal{L}_{\mathrm{om}}$ is written as:
\begin{equation}
\begin{aligned}
\mathcal{L}_{\mathrm{om}}(\hat{r}, r, y)
&= \frac{-1}{2B}\sum_{i=1}^{B}\Bigg[
\log
\frac{\sum_{j=1}^{B} M_{ij}\exp\!\left(s(\hat{r}_i,r_j)/\tau\right)}
{\sum_{k=1}^{B}\exp\!\left(s(\hat{r}_i,r_k)/\tau\right)}
\\
&+\log
\frac{\sum_{j=1}^{B} M_{ij}\exp\!\left(s(r_i,\hat{r}_j)/\tau\right)}
{\sum_{k=1}^{B}\exp\!\left(s(r_i,\hat{r}_k)/\tau\right)}
\Bigg],
\end{aligned}
\end{equation}
where, the $s$ is $cos(\hat{r},r)$, and the $\tau$ is the temperature.

\paragraph{Diffusion prior.}
Let $h_i$ be the target CLIP image embedding.
The diffusion prior follows the standard denoising objective: it receives a noisy embedding $\delta_t$, predicts $\hat{\epsilon}_\theta(\delta_t,t,c_i)$ under the EEG semantic condition $c_i$, and yields a clean image-embedding estimate $\hat{h}_i$.
The loss $\mathcal{L}_{\mathrm{noise}}$ is
\begin{equation}
\mathcal{L}_{\mathrm{noise}}
=\mathbb{E}\|\hat{\epsilon}_\theta-\epsilon\|_2^2 ,
\end{equation}
with the standard gaussian noise $\epsilon$. Finally, the \omcr{} $\mathcal{L}_{\mathrm{prior}}$ combines the denoising objective $\mathcal{L}_{\mathrm{noise}}$ with object-level contrastive loss $\mathcal{L}_{\mathrm{om}}$ and the paired cosine anchor $\mathcal{L}_{\mathrm{cos}}$:
\begin{equation}
\mathcal{L}_{\mathrm{prior}}
=\mathcal{L}_{\mathrm{noise}}
+\mathcal{L}_{\mathrm{cos}}(\hat{h}_i,h_i)
+\mathcal{L}_{\mathrm{om}}(\hat{h}_i,h_i,y).
\end{equation}

\paragraph{Low-level branch.}
Let $z_i$ be the target VAE latent and $\hat{z}_i$ the predicted latent.
The $\mathcal{L}_{\mathrm{low}}$ uses the same object-mask term as an auxiliary regularizer on top of latent regression:
\begin{equation}
\begin{aligned}
\mathcal{L}_{\mathrm{low}}
=&\|\hat{z}-z\|_1
+\lambda_{\mathrm{mse}}\|\hat{z}-z\|_2^2\\
&+\mathcal{L}_{\mathrm{cos}}(\hat{z},z)
+\lambda_o^\ell \mathcal{L}_{\mathrm{om}}(\hat{z},z,y),
\end{aligned}
\end{equation}
where, $\lambda_{\mathrm{mse}}$ and $\lambda_o^\ell$ are weights for balancing scale differences. Because VAE latents also encode layout and texture of individual images, the term $\mathcal{L}_{\mathrm{low}}$ serves only as an auxiliary constraint rather than the main \omcr{} source of object-level preservation.

\section{Experiments}

\subsection{Datasets and Metrics}

Evaluation uses \things{}, which provides dense RSVP EEG responses to 1,854 visual object concepts with a standard split of 1,654 training object concepts and 200 test object concepts
\citep{gifford2022large}.
Note that \things{} is the only publicly available dataset for this work.
For reconstruction, we follow prior brain-to-image works
\citep{scotti2023mindeye,li2024atm}
and report Alex.2, Alex.5, CLIP, and SwAV, with PixCorr (Pix.) and SSIM included in complete reconstruction tables
\citep{krizhevsky2012alexnet,caron2020swav,wang2004ssim}.
Detailed metric definitions and sanity checks are provided in the Appendix A.

\subsection{Experimental Settings}

EEG and image preprocessing follow the augmentation and semantic-projection protocol of related EEG-to-image work \citep{zhang2025neurobridge}.
Meanwhile, consistent with prior reconstruction works \cite{li2024atm,dai2026avde}, subject 08 is chosen as the primary subject for cross-subject comparison.

We implement the reconstruction pipeline using SDXL-Turbo with IP-Adapter
\citep{podell2023sdxl,ye2023ipadapter}.
Its diffusion prior builds on the U-net
with the hyperparameter set to $\tau=0.07$ \citep{li2024atm}. %
The low-level branch adopts latent regression with low-level OMCR under $\lambda_o^\ell=0.5$ and $\lambda_{\mathrm{mse}}=0.01$.
For \scgabbr{}, the CORAL teacher and anchor variants use fixed pairs
$(r,\rho)\in\{(2,0.9),(4,0.9),(6,1.0),(2,1.0),(6,0.9)\}$, covering CORAL regularization strengths ranging from relatively aggressive to mild.
The ridge regularization is configured with $\lambda=1$, while the residual branch blend is $\beta=0.125$.

All experiments are implemented in PyTorch 2.5.1 and run on a single NVIDIA RTX 4090 GPU with 24 GB.
All reported SCG reconstruction results adopt the same \omcr{} settings.
Additional setting details are provided in the Appendix B and C.

\subsection{Brain Decoding Performance}

\begin{table}[!t]
\centering
\small
\setlength{\tabcolsep}{0.5pt}
\begin{tabular}{@{}lcccccc@{}}
\toprule
Method & Pix. $\uparrow$ & SSIM $\uparrow$ & Alex.2 $\uparrow$ & Alex.5 $\uparrow$ & CLIP $\uparrow$ & SwAV $\downarrow$ \\
\midrule
\multicolumn{7}{c}{\textit{Cross-subject}} \\
\cmidrule(lr){1-7}
Li et al. & 0.079 & 0.374 & 0.640 & 0.649 & 0.556 & 0.638 \\
\reconreadout{} & \textbf{0.174} & \textbf{0.389} & \textbf{0.684} & \textbf{0.726} & \textbf{0.723} & \textbf{0.618} \\
\midrule
\multicolumn{7}{c}{\textit{Intra-subject}} \\
\cmidrule(lr){1-7}
Li et al. & 0.160 & 0.345 & 0.776 & 0.866 & 0.786 & 0.582 \\
CogCapturer & 0.175 & 0.366 & 0.760 & 0.610 & 0.744 & 0.577 \\
AVDE & \underline{0.188} & \underline{0.396} & \underline{0.817} & \underline{0.889} & \underline{0.795} & \underline{0.557} \\
\midrule
\reconreadout{} & \textbf{0.304} & \textbf{0.397} & \textbf{0.868} & \textbf{0.905} & \textbf{0.888} & \textbf{0.538} \\
\bottomrule
\end{tabular}
\caption{Subject-08 reconstruction metrics in cross- and intra-subject settings. CogCapturer abbreviates CognitionCapturer~\cite{zhang2025cognitioncapturer}.}
\label{tab:s8-main}
\end{table}

\begin{table}[!t]
\centering
\small
\setlength{\tabcolsep}{1mm}
\begin{tabular}{@{}llcccc@{}}
\toprule
Method & Stat & Alex.2 $\uparrow$ & Alex.5 $\uparrow$ & CLIP $\uparrow$ & SwAV $\downarrow$ \\
\midrule
Li et al. & Mean & 0.590 & 0.612 & 0.552 & 0.648 \\
 & Std & 0.025 & 0.021 & 0.032 & 0.006 \\
\reconreadout{} & Mean & \textbf{0.706} & \textbf{0.745} & \textbf{0.713} & \textbf{0.618} \\
 & Std & 0.028 & 0.023 & 0.027 & 0.010 \\
\bottomrule
\end{tabular}
\caption{Cross-subject reconstruction under ten-subject LOSO, reported as mean and population standard deviation.}
\label{tab:cross-loso-recon-compare}
\end{table}

Table~\ref{tab:retrieval-full} reports cross-subject 200-way retrieval under ten-subject LOSO. \retreadout{} outperforms the the strongest baseline considered, NeuroBridge, by 0.073 in Top-1 accuracy, indicating improved target-subject semantic separability. Table~\ref{tab:s8-main} evaluates Sub-08 reconstruction. In the cross-subject setting, \reconreadout{} outperforms Li et al. with gains of 0.167 on CLIP and 0.077 on Alex.5; in the intra-subject setting, it achieves the best performance across the reported semantic and perceptual metrics. Table~\ref{tab:cross-loso-recon-compare} further shows that the reconstruction advantage generalizes across subjects, with ten-subject LOSO gains over Li et al. of 0.161 on CLIP and 0.133 on Alex.5.
These results support task-specific readout selection in SPSC: retrieval benefits from higher-rank discriminative coordinates, whereas reconstruction benefits from lower-rank semantic conditions that remain more stable for diffusion-based synthesis.
Complete per-subject reconstruction and retrieval results for cross- and intra-subject settings are provided in the Appendix E and F.

\begin{figure}[ht]
\centering
\includegraphics[width=\linewidth,height=0.60\textheight,keepaspectratio]{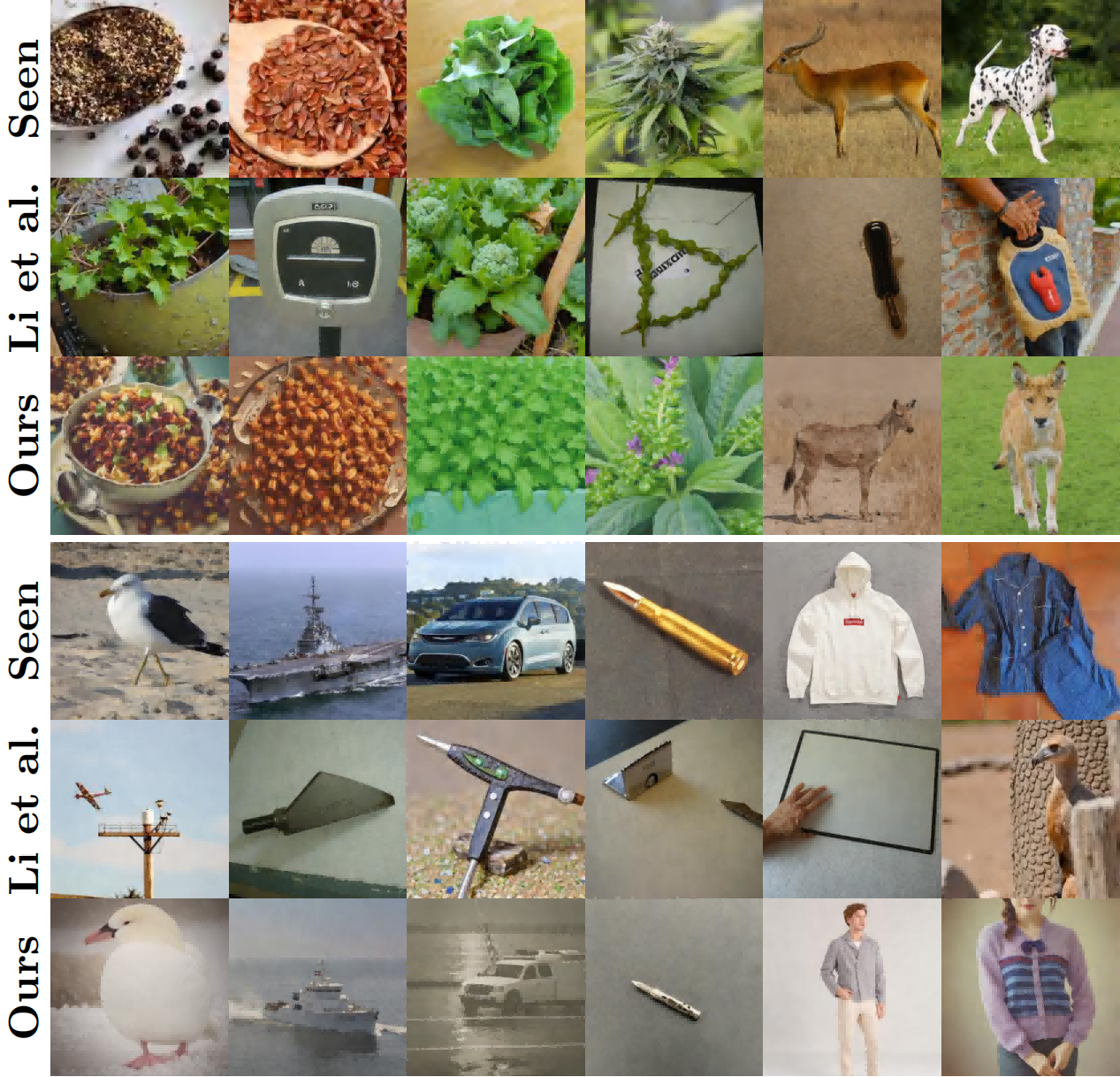}
\caption{Cross-subject Sub-08 reconstruction examples.}
\label{fig:inter-qual}
\end{figure}

Figure~\ref{fig:inter-qual} visually illustrates the same trend: calibrated low-rank semantics preserve object identity more consistently than Li et al. under the same generation backend.

\subsection{Ablation Studies}
\paragraph{Reconstruction ablation.}

\begin{table}[ht]
\centering
\small
\setlength{\tabcolsep}{2pt}
\begin{tabular}{lcccc}
\toprule
Method & Alex.2 $\uparrow$ & Alex.5 $\uparrow$ & CLIP $\uparrow$ & SwAV $\downarrow$ \\
\midrule
Base & 0.508 & 0.513 & 0.527 & 0.718 \\
Base + \omcr{} & 0.676 & 0.680 & 0.679 & 0.631 \\
+\fulldiag{} & 0.677 & 0.696 & 0.680 & 0.627 \\
+\retreadout{} & \underline{0.678} & \underline{0.702} & \underline{0.694} & \underline{0.622} \\
+\reconreadout{} & \textbf{0.684} & \textbf{0.726} & \textbf{0.723} & \textbf{0.618} \\
\bottomrule
\end{tabular}
\caption{LOSO Sub-08 reconstruction ablation.}
\label{tab:sub08-two-component}
\end{table}

Table~\ref{tab:sub08-two-component} separates backend preservation from target semantic calibration for cross-subject Sub-08.
\omcr{} delivers the primary performance gain by preserving object-level positives during backend training, increasing the CLIP by 0.152 over Base.
When \omcr{} is enabled, \scgabbr{} changes only the semantic condition fed into the prior training and inference, and \reconreadout{} yields the best reconstruction metrics and provides another 0.044 the CLIP gain.
These results validate the intended division of labor between object-level feature preservation and target-subject semantic calibration.

\begin{figure}[ht]
\centering
\includegraphics[width=\linewidth]{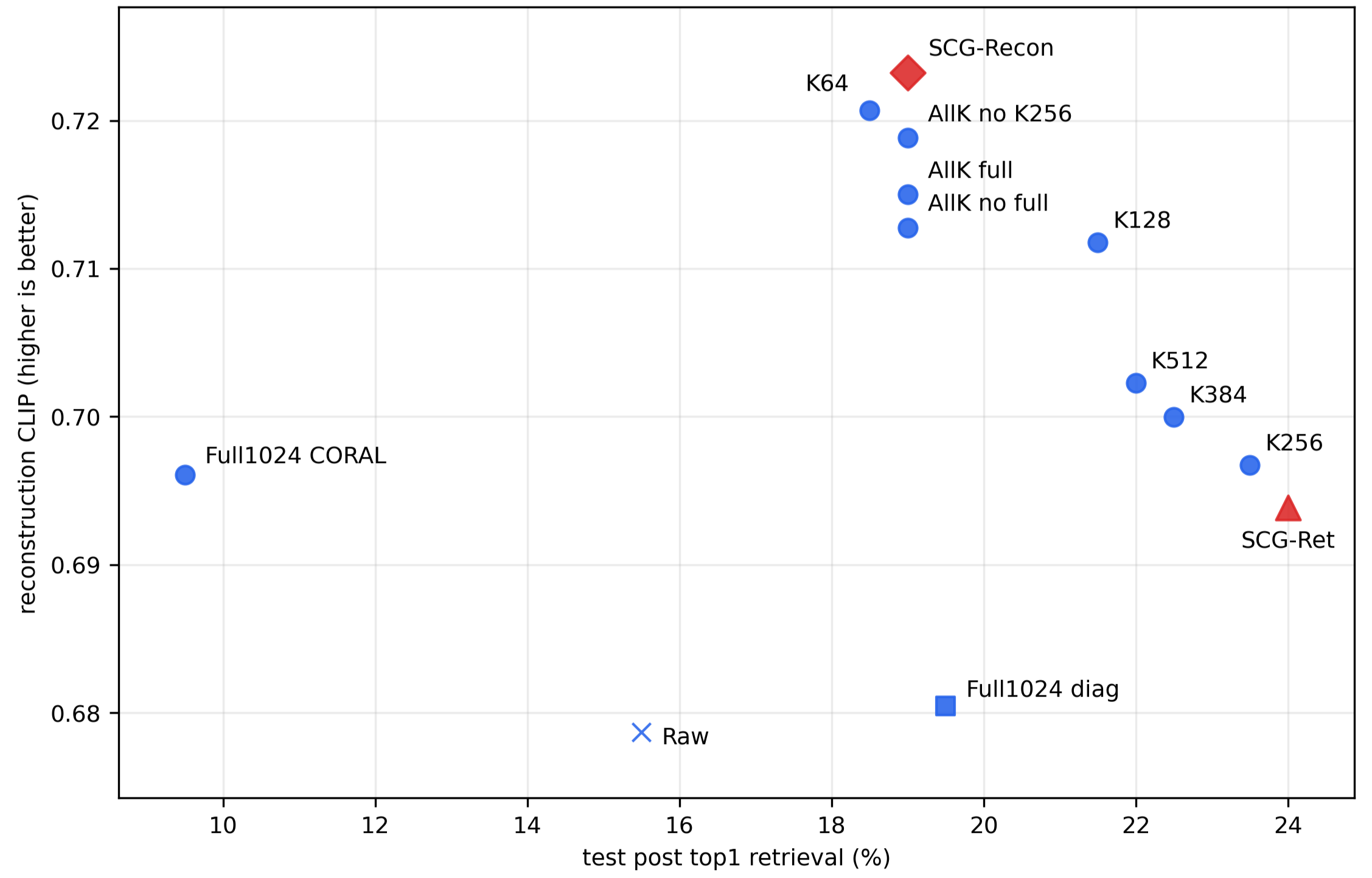}
\caption{LOSO Sub-08 retrieval--reconstruction trade-off.}
\label{fig:pareto}
\end{figure}
\paragraph{SCG candidate behavior.}

Figure~\ref{fig:pareto} shows that retrieval and reconstruction prefer different semantic scales under the same \omcr{} backend.
K256 and \retreadout{} occupy the retrieval-oriented region, while K64/K128 and \reconreadout{} give better reconstruction performance.
The two readouts therefore represent task-adapted representations derived from the same candidate semantic pool.
On the LOSO Sub-08 test, \retreadout{} reaches a Top-1 score of 24.0, whereas \reconreadout{} attains the highest reconstruction CLIP of 0.723.
Full candidate-level retrieval and reconstruction metrics are provided in the Appendix D.

\subsection{Analysis and Discussion}
\paragraph{Subspace reliability and subject shift.}

\begin{figure}[htbp]
\centering
\includegraphics[width=\linewidth]{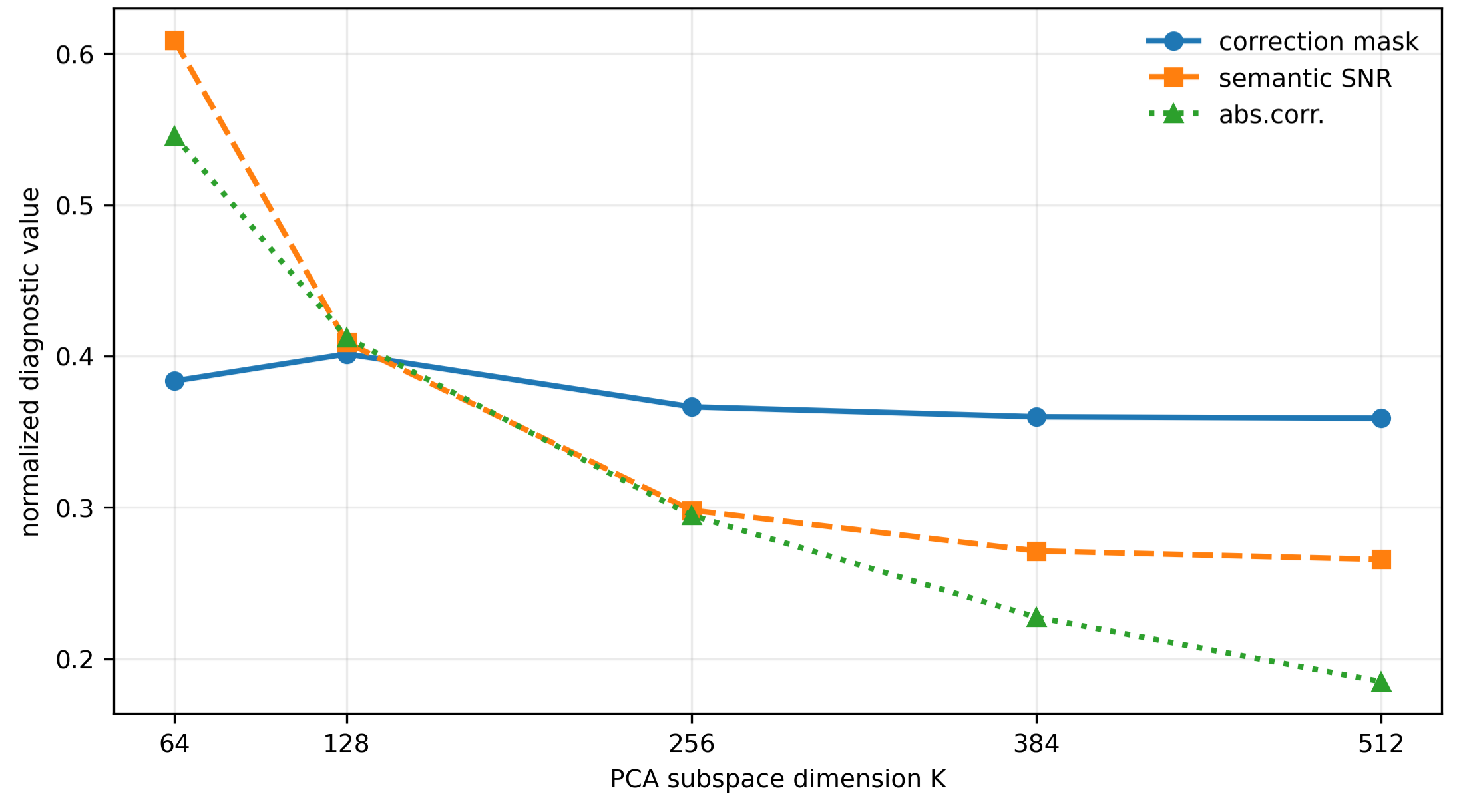}
\caption{Diagnostics for LOSO training subspace scale. The correction mask estimates the required correction magnitude for each PCA coordinate; semantic SNR is defined as the variance across stimuli divided by the variance across subjects for the same stimulus; abs.corr. denotes the mean absolute Pearson correlation between EEG embedding coordinate means of source subjects and paired image prototype coordinates.}
\label{fig:subspace-scale-analysis}
\end{figure}

Figure~\ref{fig:subspace-scale-analysis} measures correction strength, semantic SNR, and EEG--image alignment across varying image-PCA ranks.
Low ranks yield higher SNR and stronger EEG--image correlations, while higher ranks introduce finer but less stable coordinate representations.
This pattern explains why \reconreadout{} uses low ranks for generation and \retreadout{} keeps mid/high ranks for retrieval.

\paragraph{Confidence gate and selector behavior.}

\begin{table}[htbp]
\centering
\small
\setlength{\tabcolsep}{1pt}
\begin{tabular}{llcc}
\toprule
Readout & Selector & Top-1 (\%) $\uparrow$ & Top-5 (\%) $\uparrow$ \\
\midrule
\retreadout{} & Confidence & \underline{24.0} & 46.0 \\
\retreadout{} & Fixed k256 & 23.5 & \underline{46.5} \\
\retreadout{} & Average & 21.0 & 46.0 \\
\retreadout{} & Oracle & \textbf{27.0} & \textbf{47.5} \\
\bottomrule
\end{tabular}
\caption{LOSO Sub-08 retrieval selector behavior.}
\label{tab:selector-ablation}
\end{table}

The confidence gate enables label-free routing based on semantic similarity sharpness against training prototypes.
The nearest-prototype term measures manifold proximity, while the tail gap captures local sharpness.
Table~\ref{tab:selector-ablation} compares this routing rule with fixed, average, and oracle alternatives; the oracle row serves as a post-hoc diagnostic that evaluates candidate complementarity unavailable to the label-free gate.
The gate turns a single SCG candidate family into task-conditioned readouts: \retreadout{} routes toward high-rank semantics for retrieval discrimination, whereas \reconreadout{} selects smoother low-rank semantics for generation.
Additional selector diagnostics in the Appendix D show the routing behavior for \reconreadout{}.

\paragraph{Semantic alignment of the image side.}

\begin{figure}[htbp]
\centering
\includegraphics[width=\linewidth]{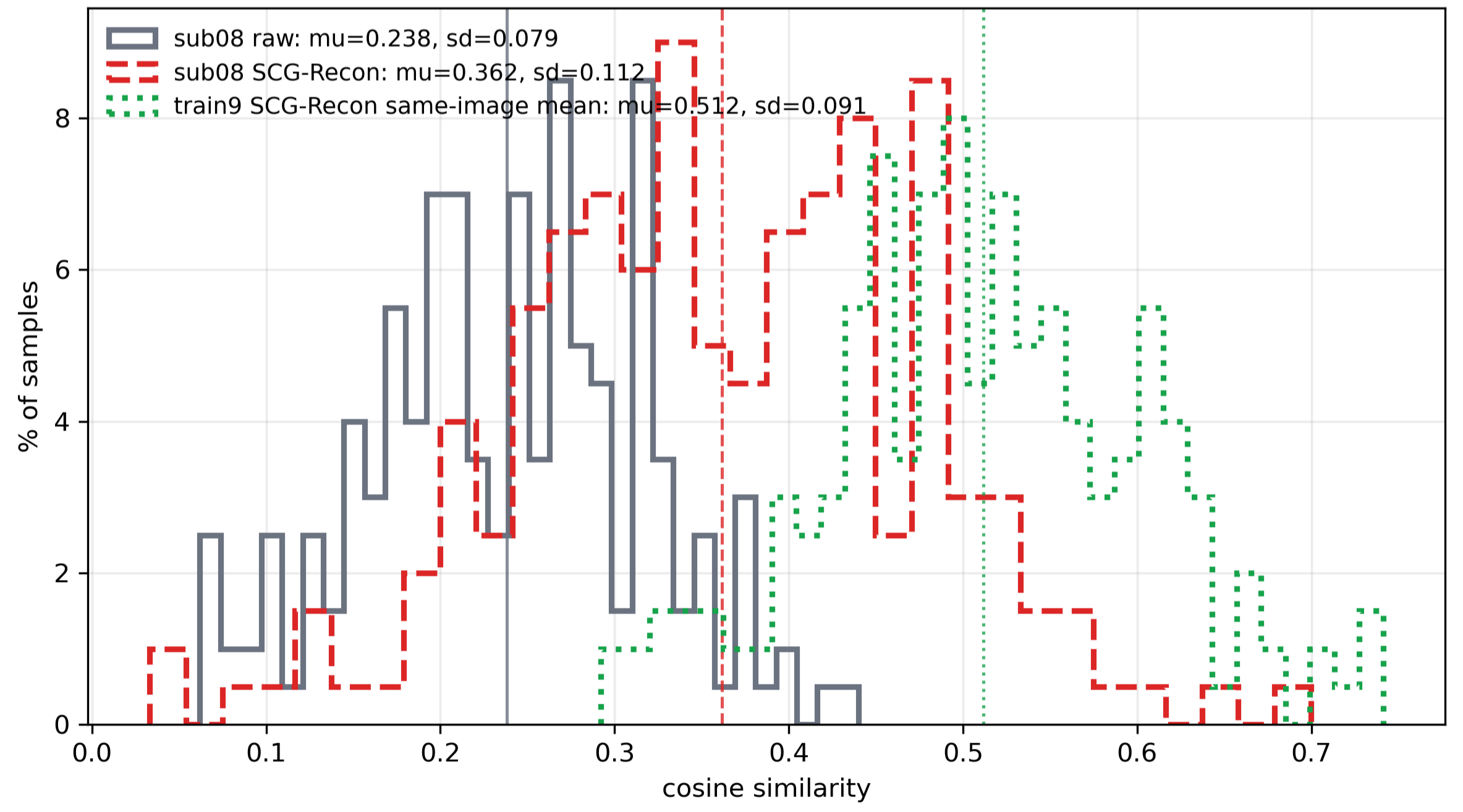}
\caption{LOSO Sub-08 paired semantic alignment.}
\label{fig:image-side-alignment}
\end{figure}

Figure~\ref{fig:image-side-alignment} relies solely on held-out image embeddings after inference to measure whether the exported semantic condition moves toward the matched test embedding.
For each test stimulus, the score corresponds to the cosine similarity between the corrected EEG semantic and the semantic prototype of the test image.
Raw Sub-08 semantics generate a mean cosine score of 0.238, while \reconreadout{} shifts the target semantics up to 0.362; averaging responses from the 9 training source subjects for the same image reaches 0.512, which acts as an analytical reference for image-specific cross-subject alignment.
This comparison with paired images links the reconstruction readout to generator conditioning: it supplies a smoother semantic condition for the target subject that lies closer to the image subspace within the shared embedding space.

\paragraph{Cross-subject prior analysis.}

\begin{figure}[htbp]
\centering
\includegraphics[width=\linewidth]{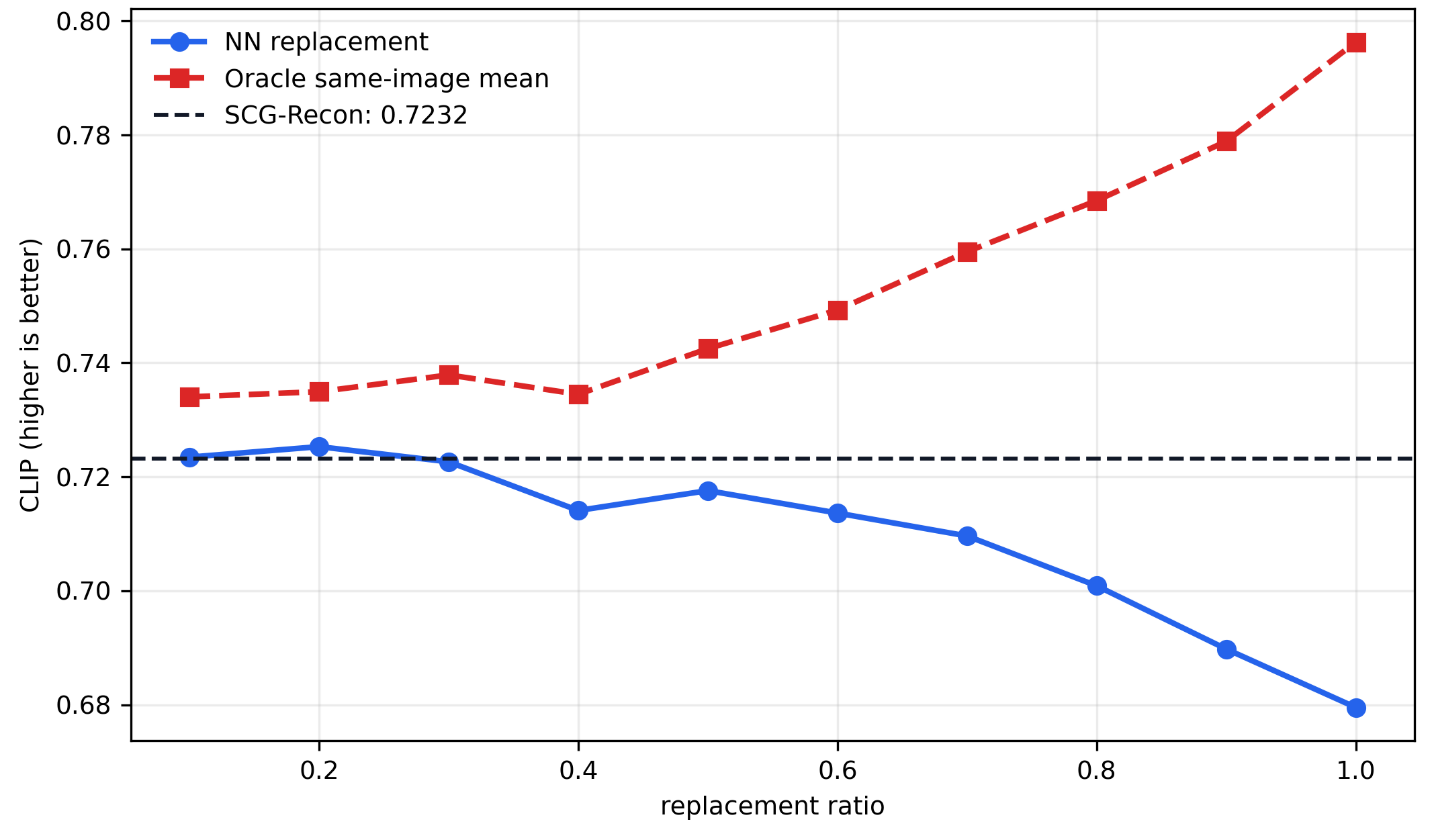}
\caption{LOSO Sub-08 prior replacement curve.}
\label{fig:prior-replacement-curve}
\end{figure}

Figure~\ref{fig:prior-replacement-curve} examines whether source-subject test responses contain reusable image-specific structure.
Nearest-neighbor (NN) replacement operates in a label-free way but risks overwriting the target manifold geometry, while oracle replacement uses same-image identity and serves only as an upper-bound reference.
NN replacement delivers only marginal gains at low replacement ratios and then underperforms \reconreadout{} as more target semantics get overwritten. In contrast, the oracle curve rises with greater access to the test identities.
\reconreadout{} remains practically deployable since it calibrates the target response itself rather than replacing it with source-subject semantics.

\section{Conclusion}

Our SPSC framework addresses zero-shot cross-subject EEG-to-image reconstruction as semantic conditioning for generative models.
Specifically, \omcr{} preserves object-level neighborhoods during generation backend training, and \scgabbr{} calibrates the target subject's EEG semantics after the projector using training image prototypes and unlabeled target EEG statistics.
Under ten-subject LOSO evaluation, the reconstruction readout improves key semantic reconstruction metrics during label-free inference for the target subject.
The analyses show that low-rank calibrated semantics better align the target EEG condition with the embedding manifold of the image side for reconstruction, while higher-rank semantics retain discriminative structure for retrieval.

{
\bibliography{references}
}

\clearpage
\appendix
\twocolumn[
\centering
{\Large\bf Appendix\par}
{\large Supplemental Material for ``SPSC: Zero-Shot Cross-Subject EEG-to-Image Reconstruction''\par}
]

\section{A. Metric Definitions and Sanity Check}
\label{sec:metric-sanity}

Following prior brain-to-image evaluation protocols, Alex.2 and Alex.5 denote two-way identification scores computed from AlexNet layers 2 and 5, and CLIP denotes the corresponding identification score computed from CLIP embeddings. %
SwAV is the feature distance from SwAV-ResNet50, with lower values indicating stronger feature similarity. %
PixCorr and SSIM quantify low-level image similarity and are reported as auxiliary reconstruction measures. %

\begin{table}[htbp]
\centering
\small
\setlength{\tabcolsep}{0.5pt}
\begin{tabular}{lcccccc}
\toprule
Condition & PixCorr $\uparrow$ & SSIM $\uparrow$ & Alex.2 $\uparrow$ & Alex.5 $\uparrow$ & CLIP $\uparrow$ & SwAV $\downarrow$ \\
\midrule
raw image & \underline{0.199} & \underline{0.284} & \underline{0.858} & \underline{0.949} & \underline{0.995} & \textbf{0.423} \\
augmentation & \textbf{0.231} & \textbf{0.293} & \textbf{0.868} & \textbf{0.953} & \textbf{0.996} & \underline{0.443} \\
\bottomrule
\end{tabular}
\caption{Oracle CLIP-to-SDXL-Turbo sanity check for ViT-H-14 embedding.}
\label{tab:oracle-clip-generator}
\end{table}

The oracle control passes ground-truth ViT-H-14 CLIP embeddings of test images directly to SDXL-Turbo with IP-Adapter, without using EEG.
This control confirms that true CLIP image embeddings yield strong semantic scores even when PixCorr and SSIM remain modest, as the generator can preserve object identity while varying texture or layout.
Accordingly, PixCorr and SSIM are treated as auxiliary reconstruction measures.

\section{B. SCG Implementation Details}
\label{sec:scg-implementation}

This section gives the implementation details condensed in the main method while preserving the same notation.
For compact notation in this appendix, define $u^K_{i,s}=u_K(x_{i,s})$, $v^K_i=u_K(p_i)$, $U_s^K=\{u^K_{i,s}\}_{i=1}^{N_{\mathrm{tr}}}$, and $V^K=\{v_i^K\}_{i=1}^{N_{\mathrm{tr}}}$.

\paragraph{Teacher variant selection.}
For each source sample, CORAL variants produce teacher coordinates $t^{K,r,\rho}_{i,s}=T_{r,\rho}(u^K_{i,s};U_s^K,V^K)$.
\scgabbr{} keeps the teacher whose decoded vector is closest to the training prototype bank:
\begin{equation}
\begin{aligned}
(r^*,\rho^*) &= \arg\max_{r,\rho}\,
\max_{p\in P}\,
\cos\!\left(d_K(t^{K,r,\rho}_{i,s}),p\right),\\
\tilde{t}^K_{i,s} &= t^{K,r^*,\rho^*}_{i,s}.
\end{aligned}
\end{equation}
The selected teacher is anchored to the paired image coordinate by the mix with a weak weight $\alpha$
\begin{equation}
q^K_{i,s}=(1-\alpha)\tilde{t}^K_{i,s}+\alpha v^K_i,\qquad \alpha=0.10 .
\end{equation}

\paragraph{Ridge feature construction.}
The ridge input concatenates the normalized coordinate, source-subject mean, log-scale, and a bias term:
\begin{equation}
\phi_K(u^K_{i,s})=
\left[
\frac{u^K_{i,s}-\mu_s^K}{\sigma_s^K},
\mu_s^K,
\log\sigma_s^K,
1
\right].
\end{equation}
Let $\Phi_K$ stack these features over source subjects, and let $Q_K$ stack the teacher coordinates $q^K_{i,s}$.
The closed-form ridge solution is
\begin{equation}
W_K=(\Phi_K^\top\Phi_K+\lambda I)^{-1}\Phi_K^\top Q_K,\qquad \lambda=1 .
\end{equation}
With $S$ source subjects, $\Phi_K\in\mathbb{R}^{(N_{\mathrm{tr}}S)\times(3K+1)}$, $Q_K\in\mathbb{R}^{(N_{\mathrm{tr}}S)\times K}$, and $W_K\in\mathbb{R}^{(3K+1)\times K}$.

\paragraph{Target anchor selection.}
For the target batch, \scgabbr{} keeps the anchor variant with the largest prototype-bank cosine:
\begin{equation}
\begin{aligned}
(r_K^*,\rho_K^*) &= \arg\max_{r,\rho}\max_{p\in P}\cos(a_K^{r,\rho}(x),p),\\
a_K(x)&=a_K^{r_K^*,\rho_K^*}(x).
\end{aligned}
\end{equation}
The final readouts use the same candidate family as in the main text, with \retreadout{} selecting from $\{0,256,384,512\}$ and \reconreadout{} selecting from $\{0,64,128\}$.

\section{C. Intra-Subject SCG Setting and Details}

Intra-subject experiments evaluate the backend and calibration family under reduced subject shift.

\subsection{SCG in the Intra-Subject Setting}
\label{sec:intra-scg-setting}

When training and test samples come from the same subject, the intra-subject variant uses the same calibration set with statistics from a single subject.
The image prototype bank and PCA bases are unchanged.
For a training stimulus $i$, let
\begin{equation}
u_i^K = B_K^\top(x_i-\mu_P),\qquad
v_i^K = B_K^\top(p_i-\mu_P)
\end{equation}
be the EEG coordinates and image prototype coordinates in a $K$-dimensional PCA subspace of image prototypes.
In LOSO, the correction mask is informed by variation across subjects for the same stimulus.
In the intra-subject branch, this term is replaced by the coordinate-wise EEG--image distribution gap
\begin{equation}
d^K =
\left|\frac{\hat{\mu}_X^K-\hat{\mu}_P^K}{\hat{\sigma}_P^K}\right|
+
\left|\log\frac{\hat{\sigma}_X^K}{\hat{\sigma}_P^K}\right|,
\end{equation}
where $\hat{\mu}_X^K,\hat{\sigma}_X^K$ are training EEG coordinate statistics and $\hat{\mu}_P^K,\hat{\sigma}_P^K$ are training statistics of image prototype coordinates.
Coordinates with larger $d^K$ are more eligible for calibration, while coordinates with high variance across images or high EEG--image coordinate correlation are protected as semantic axes.
The same candidate construction and readout based on prototype confidence are then used at test time with unlabeled test EEG batch statistics.

\subsection{Intra-Subject Subject-08 Decoding Performance}

\begin{figure}[!t]
\centering
\includegraphics[width=\linewidth,height=0.60\textheight,keepaspectratio]{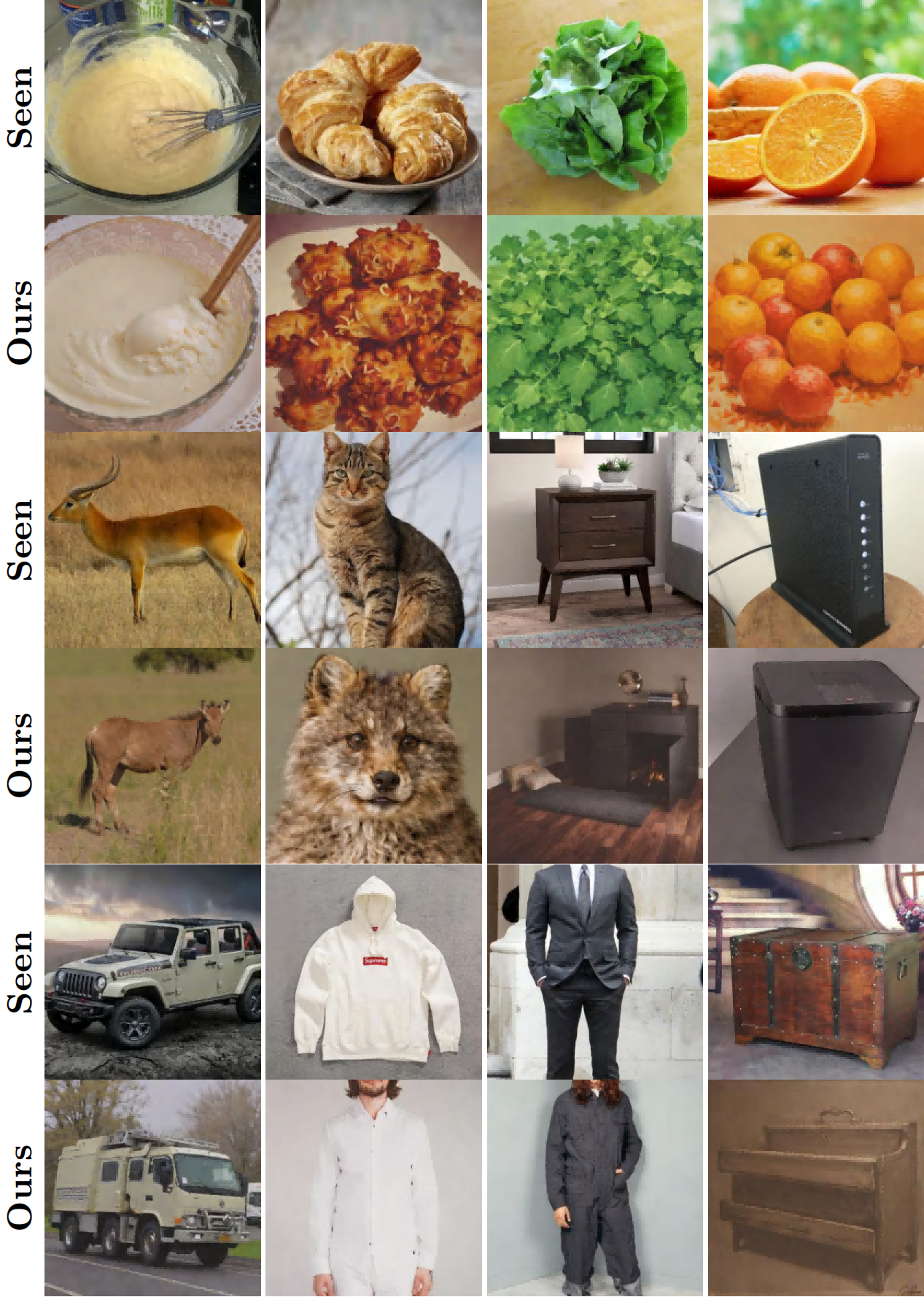}
\caption{Intra-subject Subject-08 reconstruction examples.}
\label{fig:intra-decoding-img}
\end{figure}

Figure~\ref{fig:intra-decoding-img} presents qualitative examples for the Subject-08 intra-subject setting.

\subsection{Intra-Subject Ablations}

Table~\ref{tab:branch} attributes the intra-subject gain primarily to prior-side \omcr{}: adding this term increases the semantic metrics more substantially than low-level \omcr{}, consistent with the direct role of the predicted image embedding in IP-Adapter conditioning.
\reconreadout{} further improves CLIP while keeping the AlexNet scores and SwAV distance close to the strongest backend values, indicating that prototype-based calibration continues to refine the semantic condition even under reduced subject shift.

\begin{table}[!t]
\centering
\small
\setlength{\tabcolsep}{1.5pt}
\begin{tabular}{@{}lcccc@{}}
\toprule
Variant & Alex.2 $\uparrow$ & Alex.5 $\uparrow$ & CLIP $\uparrow$ & SwAV $\downarrow$ \\
\midrule
Base & 0.724 & 0.689 & 0.576 & 0.676 \\
+ \omcr{}(low-level) & 0.782 & 0.740 & 0.599 & 0.667 \\
+ \omcr{}(prior) & 0.852 & 0.899 & 0.861 & 0.540 \\
+ \omcr{}(current) & \underline{0.867} & \textbf{0.916} & \underline{0.866} & \textbf{0.528} \\
\reconreadout{} & \textbf{0.868} & \underline{0.905} & \textbf{0.888} & \underline{0.538} \\
\bottomrule
\end{tabular}
\caption{Intra-subject Subject-08 ablation.}
\label{tab:branch}
\end{table}

\begin{table}[!t]
\centering
\small
\setlength{\tabcolsep}{1.5pt}
\begin{tabular}{@{}lcccc@{}}
\toprule
Prior term & Alex.2 $\uparrow$ & Alex.5 $\uparrow$ & CLIP $\uparrow$ & SwAV $\downarrow$ \\
\midrule
Cosine only & 0.830 & \underline{0.892} & 0.825 & 0.580 \\
Mask contrast only & \underline{0.855} & 0.888 & \underline{0.843} & \underline{0.550} \\
Mask contrast + cosine & \textbf{0.867} & \textbf{0.916} & \textbf{0.866} & \textbf{0.528} \\
\bottomrule
\end{tabular}
\caption{Intra-subject Subject-08 regularizer ablation in the prior branch for \omcr{}.}
\label{tab:prior}
\end{table}

Table~\ref{tab:prior} further decomposes the prior-side objective.
Mask contrast improves semantic reconstruction and SwAV relative to cosine-only training, while the cosine anchor stabilizes the paired image target; combining both terms yields the strongest overall prior branch.

\subsection{Representation Analysis}

\begin{figure}[!t]
\centering
\begin{tabular}{@{}cc@{}}
\includegraphics[width=0.460\linewidth]{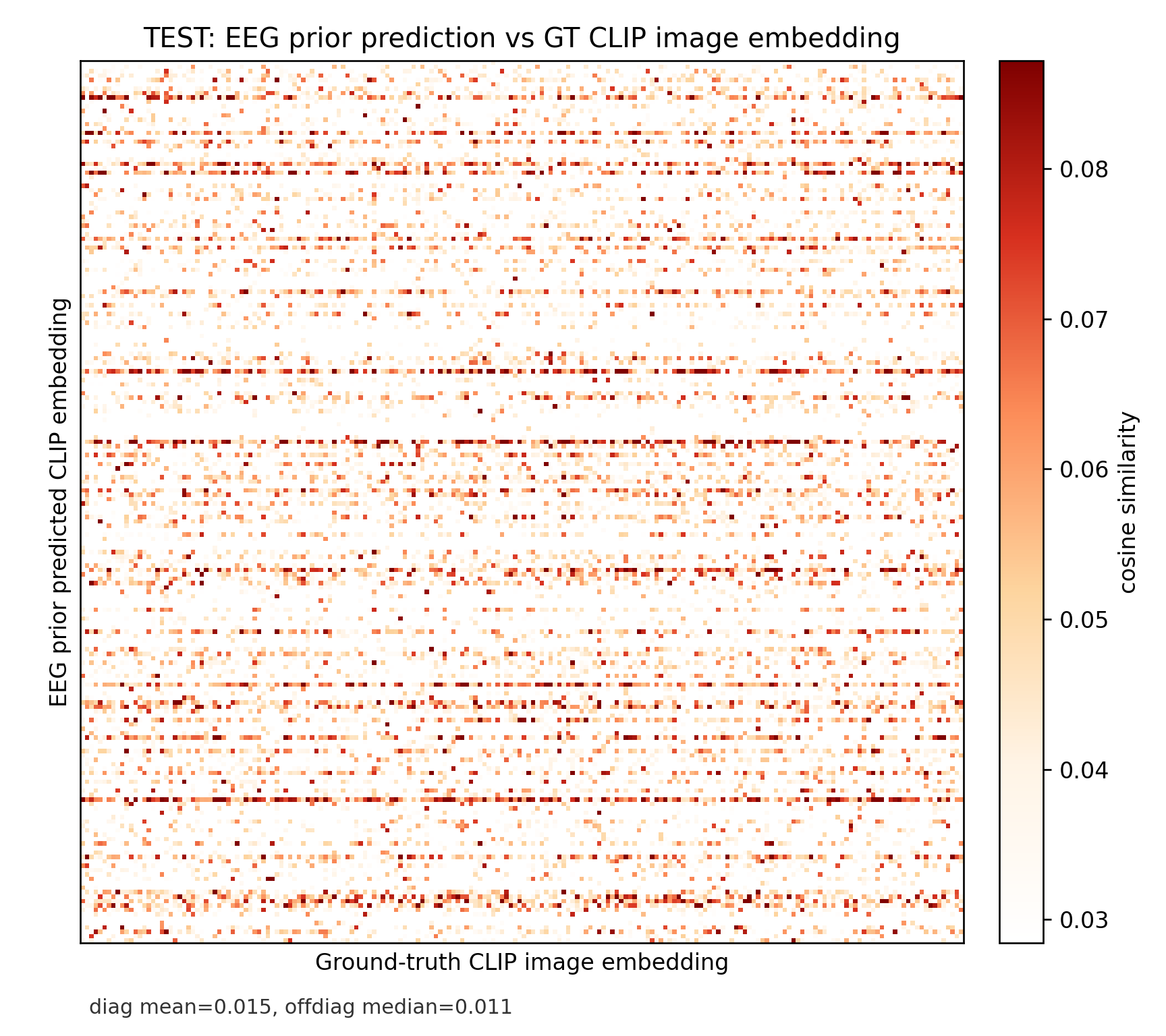} &
\includegraphics[width=0.460\linewidth]{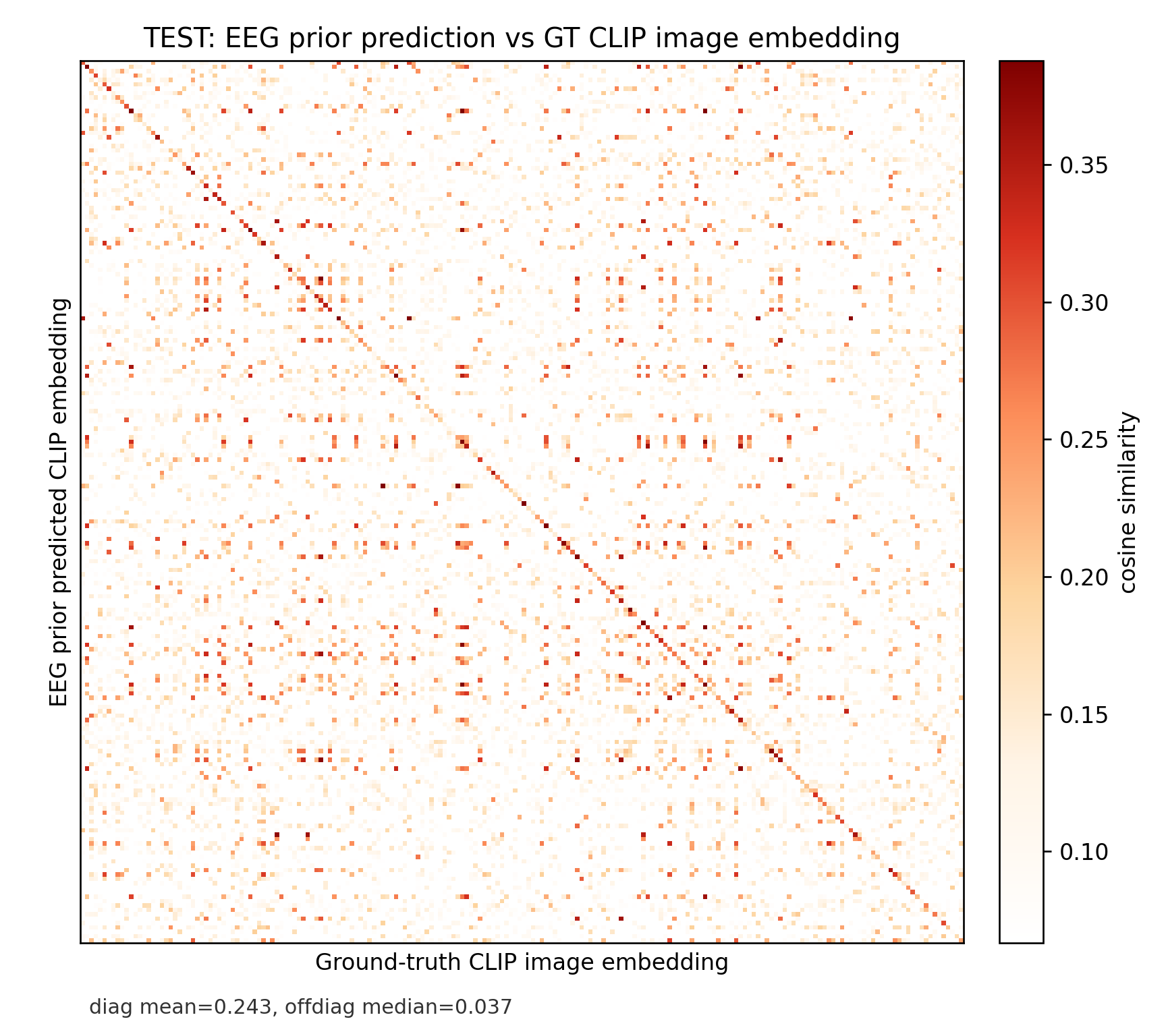} \\
\footnotesize Base & \footnotesize \reconreadout
\end{tabular}
\caption{Intra-subject Subject-08 paired similarity.}
\label{fig:rep-sim}
\end{figure}

Figure~\ref{fig:rep-sim} summarizes the representation change: after \reconreadout{}, the paired similarity matrix exhibits a clearer diagonal and weaker off-diagonal background.

\section{D. Cross-Subject SCG Diagnostics}

\subsection{Candidate Readout Trade-off}

\begin{figure}[!t]
\centering
\includegraphics[width=\linewidth]{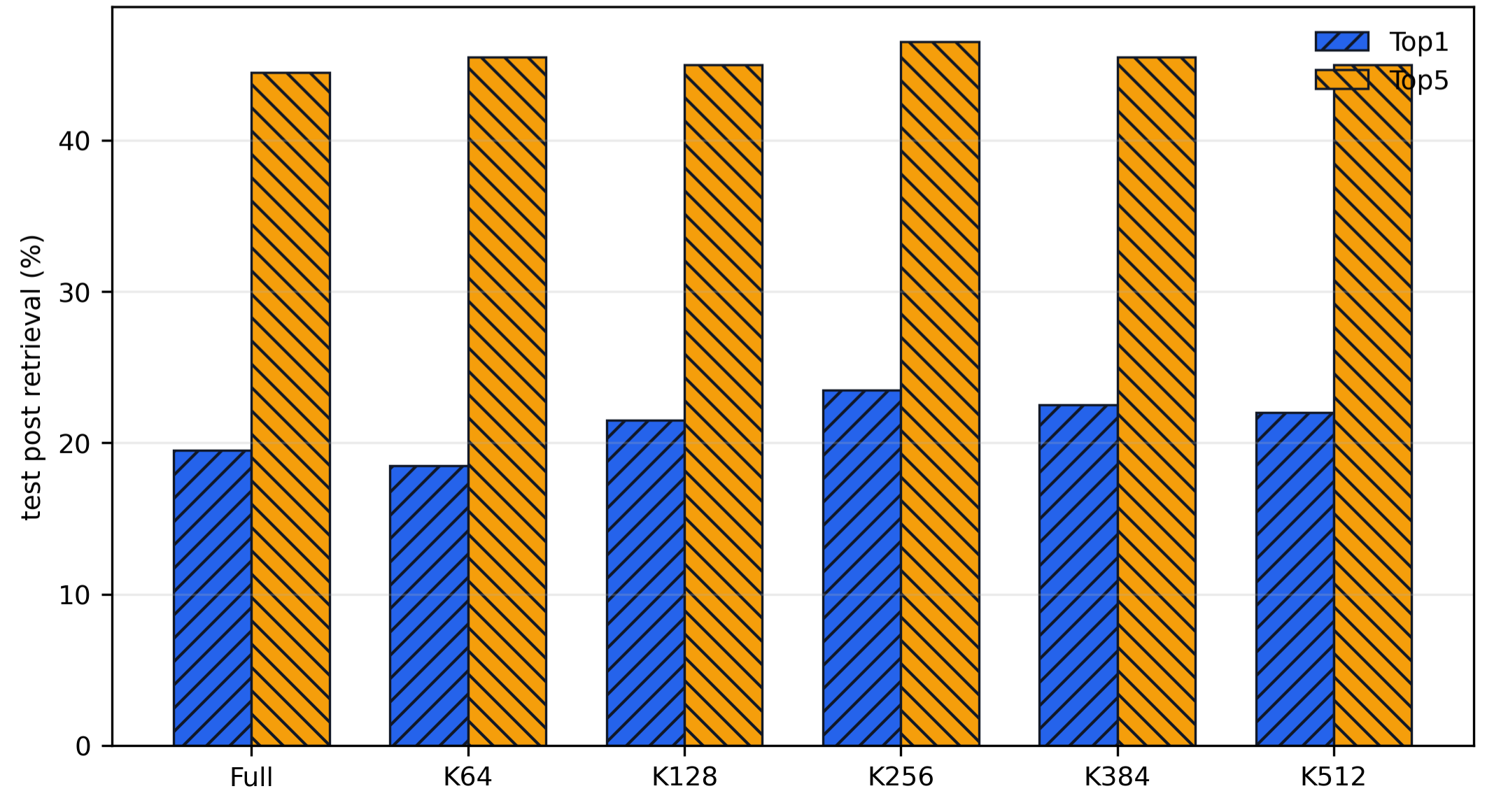}
\caption{Cross-subject Sub-08 candidate retrieval strength.}
\label{fig:candidate-strength}
\end{figure}

\begin{table}[!t]
\centering
\small
\setlength{\tabcolsep}{3pt}
\begin{tabular}{lcc}
\toprule
Case & Top-1 (\%) $\uparrow$ & Top-5 (\%) $\uparrow$ \\
\midrule
Base & 15.5 & 36.0 \\
\fulldiag{} & 19.5 & 44.5 \\
FullCORAL & 9.50 & 30.5 \\
k64 & 18.5 & 45.5 \\
k128 & 21.5 & 45.0 \\
k256 & \underline{23.5} & \textbf{46.5} \\
k384 & 22.5 & 45.5 \\
k512 & 22.0 & 45.0 \\
All k-ranks & 19.0 & 45.5 \\
\retreadout{} & \textbf{24.0} & \underline{46.0} \\
\bottomrule
\end{tabular}
\caption{Cross-subject Sub-08 SCG retrieval ablation.}
\label{tab:sub08-scg-retrieval}
\end{table}

\begin{table}[!t]
\centering
\small
\setlength{\tabcolsep}{1.2pt}
\begin{tabular}{lcccc}
\toprule
Case & CLIP $\uparrow$ & Alex.2 $\uparrow$ & Alex.5 $\uparrow$ & SwAV $\downarrow$ \\
\midrule
Base+\omcr{} & 0.679 & 0.676 & 0.680 & 0.631 \\
\fulldiag{} & 0.680 & 0.677 & 0.696 & 0.627 \\
FullCORAL & 0.696 & 0.681 & 0.711 & 0.632 \\
k64 & \underline{0.721} & 0.690 & \textbf{0.731} & \underline{0.617} \\
k128 & 0.712 & 0.691 & 0.727 & \textbf{0.617} \\
k256 & 0.697 & \textbf{0.692} & 0.699 & 0.620 \\
k384 & 0.700 & 0.685 & 0.720 & 0.622 \\
k512 & 0.702 & \underline{0.692} & 0.713 & 0.622 \\
All k-ranks & 0.715 & 0.685 & \underline{0.729} & 0.618 \\
\reconreadout{} & \textbf{0.723} & 0.684 & 0.726 & 0.618 \\
\bottomrule
\end{tabular}
\caption{Cross-subject Sub-08 SCG reconstruction ablation.}
\label{tab:sub08-scg-recon}
\end{table}

Figure~\ref{fig:candidate-strength} and Tables~\ref{tab:sub08-scg-retrieval}--\ref{tab:sub08-scg-recon} provide the candidate-level evidence for the readout split: K256 and \retreadout{} favor retrieval, whereas K64/K128 and \reconreadout{} favor reconstruction.

\subsection{Subspace and Selector Diagnostics}

\begin{figure}[!t]
\centering
\includegraphics[width=\linewidth]{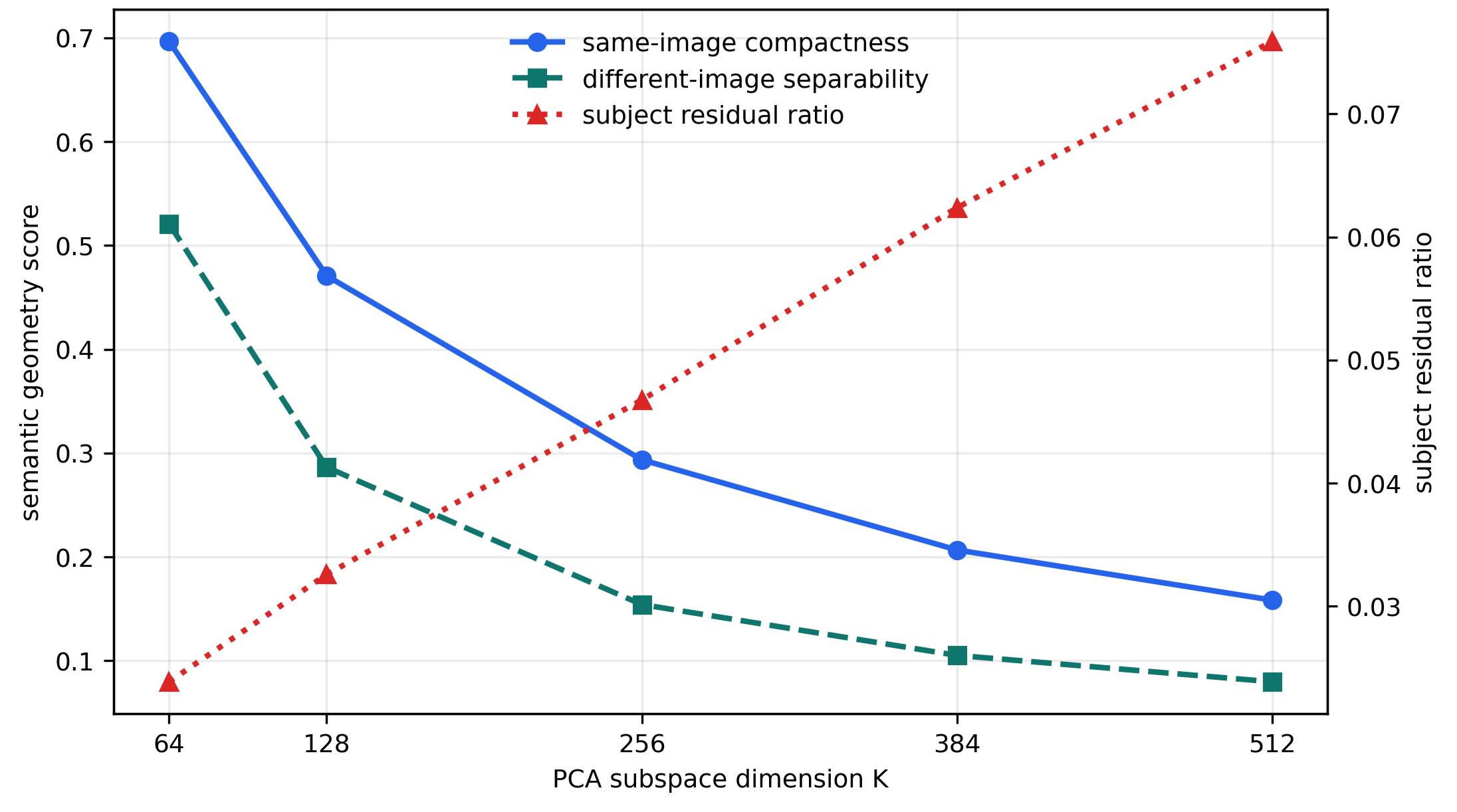}
\caption{Cross-subject Sub-08 subspace reliability.}
\label{fig:subspace-reliability}
\end{figure}

\begin{figure}[!t]
\centering
\includegraphics[width=\linewidth]{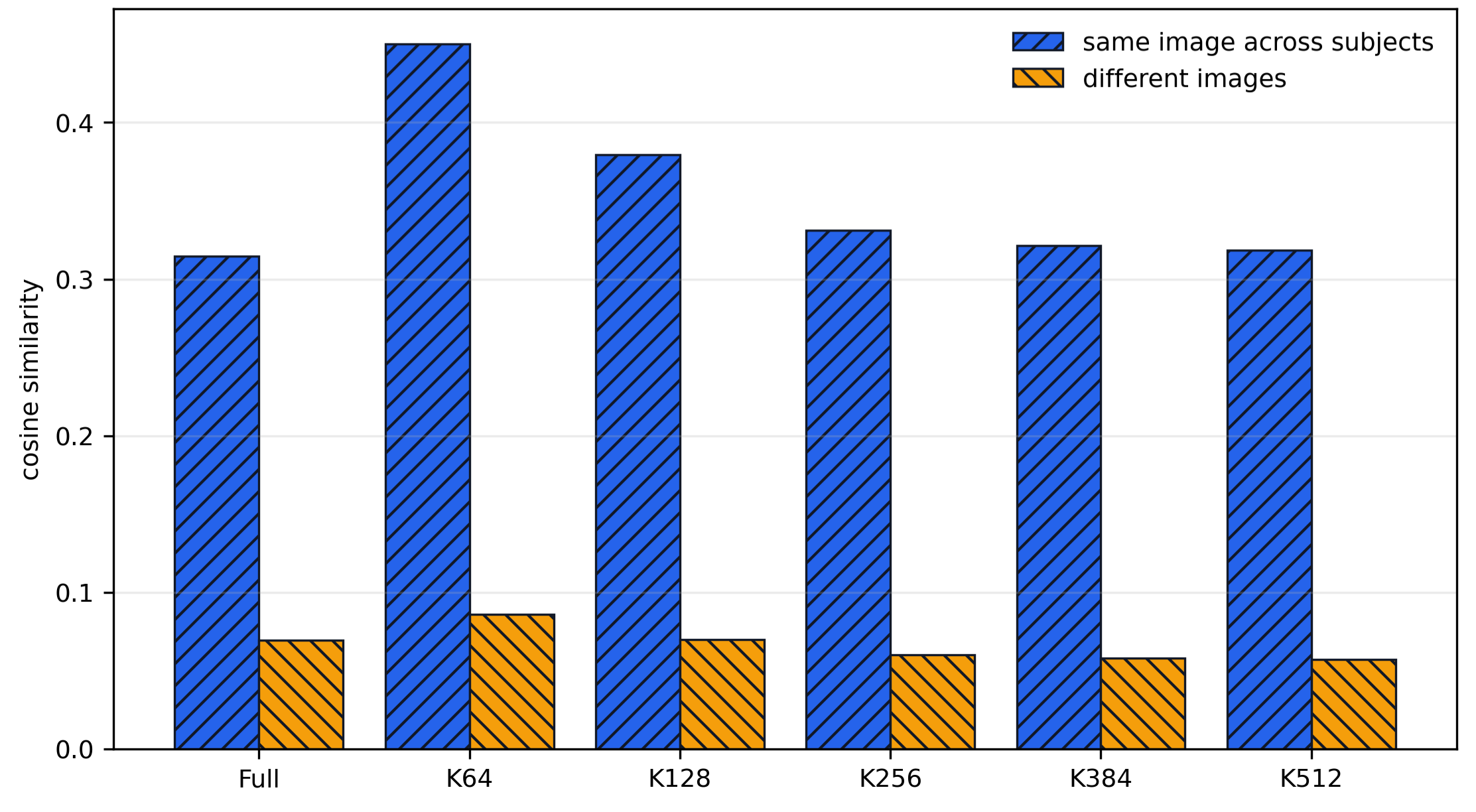}
\caption{Cross-subject Sub-08 compactness for the same stimulus.}
\label{fig:subject-shift}
\end{figure}

These diagnostics examine how the rank split relates to cross-subject semantic structure.
Figures~\ref{fig:subspace-reliability} and~\ref{fig:subject-shift} provide additional SCG subspace analyses referenced in the main text.
In Figure~\ref{fig:subspace-reliability}, compactness among responses to the same image and separability between different images are both strongest at low rank, whereas the subject residual ratio increases with the PCA dimension.
This pattern indicates that higher-rank coordinates introduce subject-sensitive residual variation more rapidly than stable cross-subject semantic structure.
Figure~\ref{fig:subject-shift} presents the same tendency in pairwise cosine geometry: responses to the same image across subjects remain substantially more similar than responses to different images, with the strongest compactness appearing at the low-rank candidate used by \reconreadout{}.

\begin{table}[!t]
\centering
\small
\setlength{\tabcolsep}{1.5pt}
\begin{tabular}{llcc}
\toprule
Readout & Selector & Top-1 (\%) $\uparrow$ & Top-5 (\%) $\uparrow$ \\
\midrule
\reconreadout{} & Confidence & 19.0 & \underline{45.5} \\
\reconreadout{} & Fixed k64 & 18.5 & \underline{45.5} \\
\reconreadout{} & Average & \underline{19.5} & \underline{45.5} \\
\reconreadout{} & Oracle & \textbf{27.0} & \textbf{49.0} \\
\bottomrule
\end{tabular}
\caption{Cross-subject Sub-08 reconstruction selector ablation.}
\label{tab:selector-ablation-recon}
\end{table}

\begin{figure}[!t]
\centering
\includegraphics[width=\linewidth]{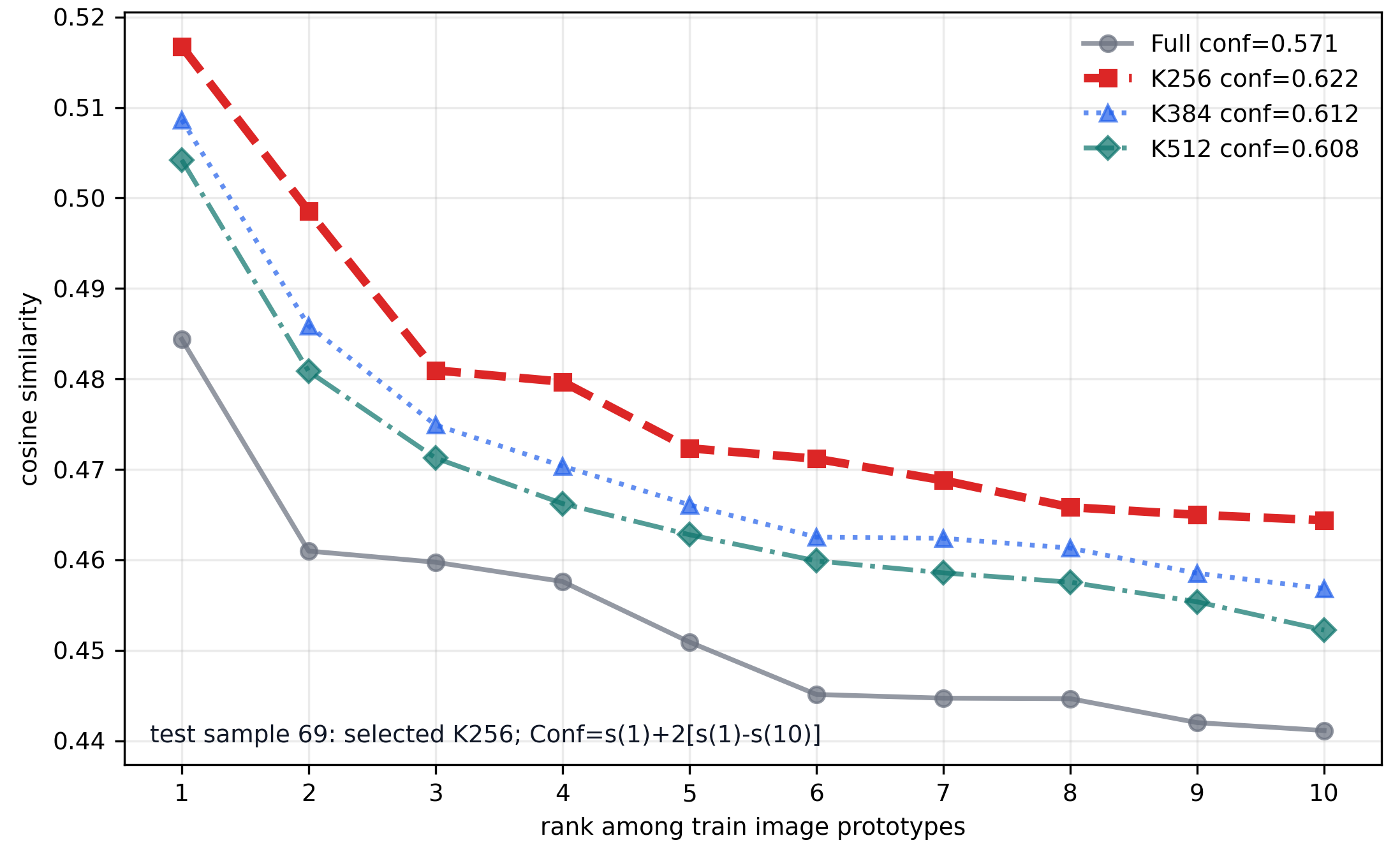}
\caption{Cross-subject Sub-08 label-free confidence selector.}
\label{fig:confidence-selector}
\end{figure}

Selector diagnostics evaluate prototype-bank confidence as a sample-wise routing rule for complete semantic candidates.
Table~\ref{tab:selector-ablation-recon} reports the low-rank \reconreadout{} selector ablation.
Fixed uses the best single candidate in the reconstruction readout set, Average uniformly averages eligible candidates, Oracle selects the best candidate by test rank for each sample after evaluation, and Confidence denotes the label-free selector used at inference.
Figure~\ref{fig:confidence-selector} visualizes the Confidence computation.
For one EEG sample from the target subject, each curve reports the top ten cosine similarities between a candidate semantic vector and the training prototype bank $P$.
The default max-tail10 score is $\operatorname{Conf}=s_{(1)}+2(s_{(1)}-s_{(10)})$, where $s_{(1)}$ and $s_{(10)}$ are the largest and tenth-largest prototype similarities.
This score combines manifold proximity and local sharpness using the training prototype bank.

\begin{figure}[!t]
\centering
\includegraphics[width=\linewidth]{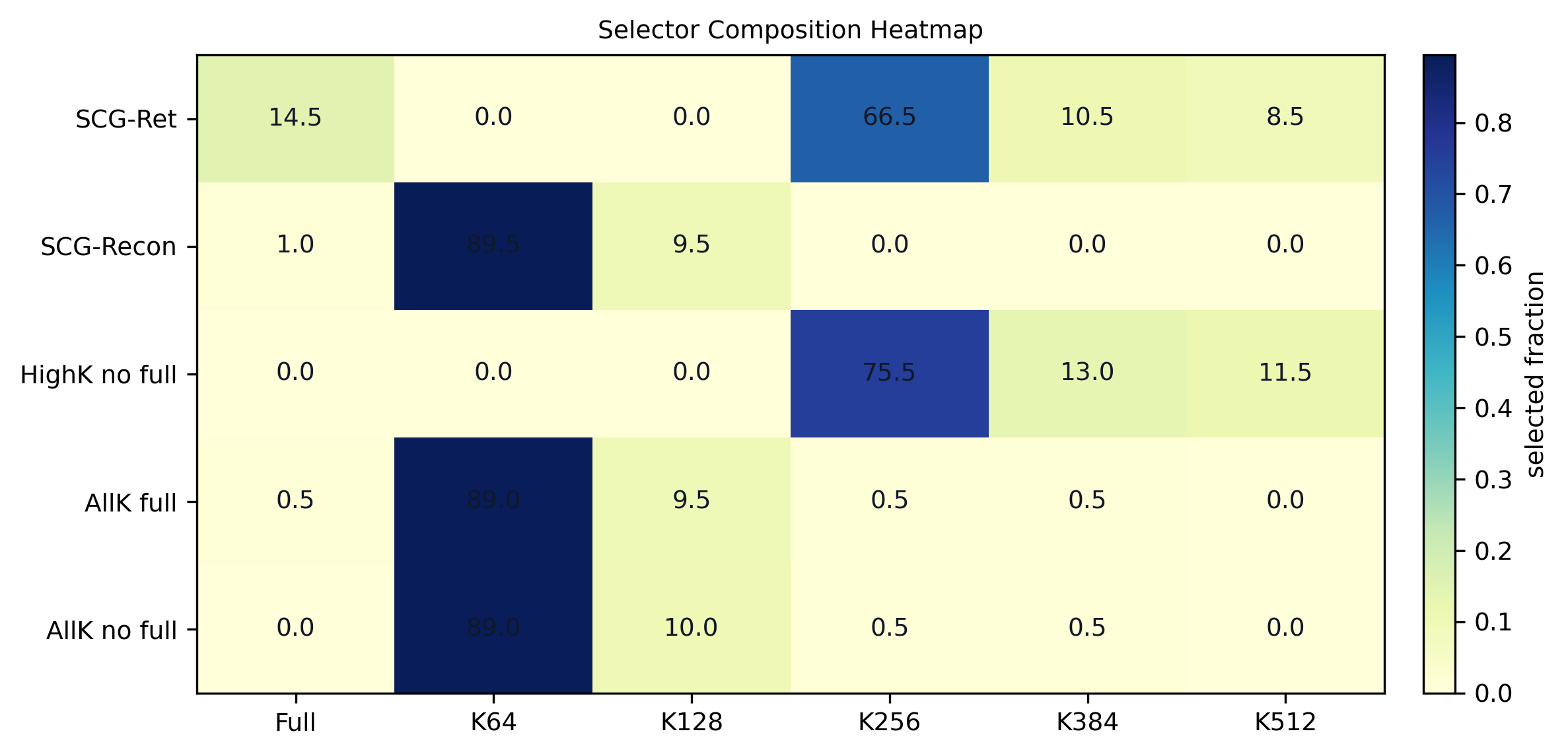}
\caption{Cross-subject Sub-08 selector composition.}
\label{fig:selector-composition}
\end{figure}

\begin{figure}[!t]
\centering
\includegraphics[width=\linewidth]{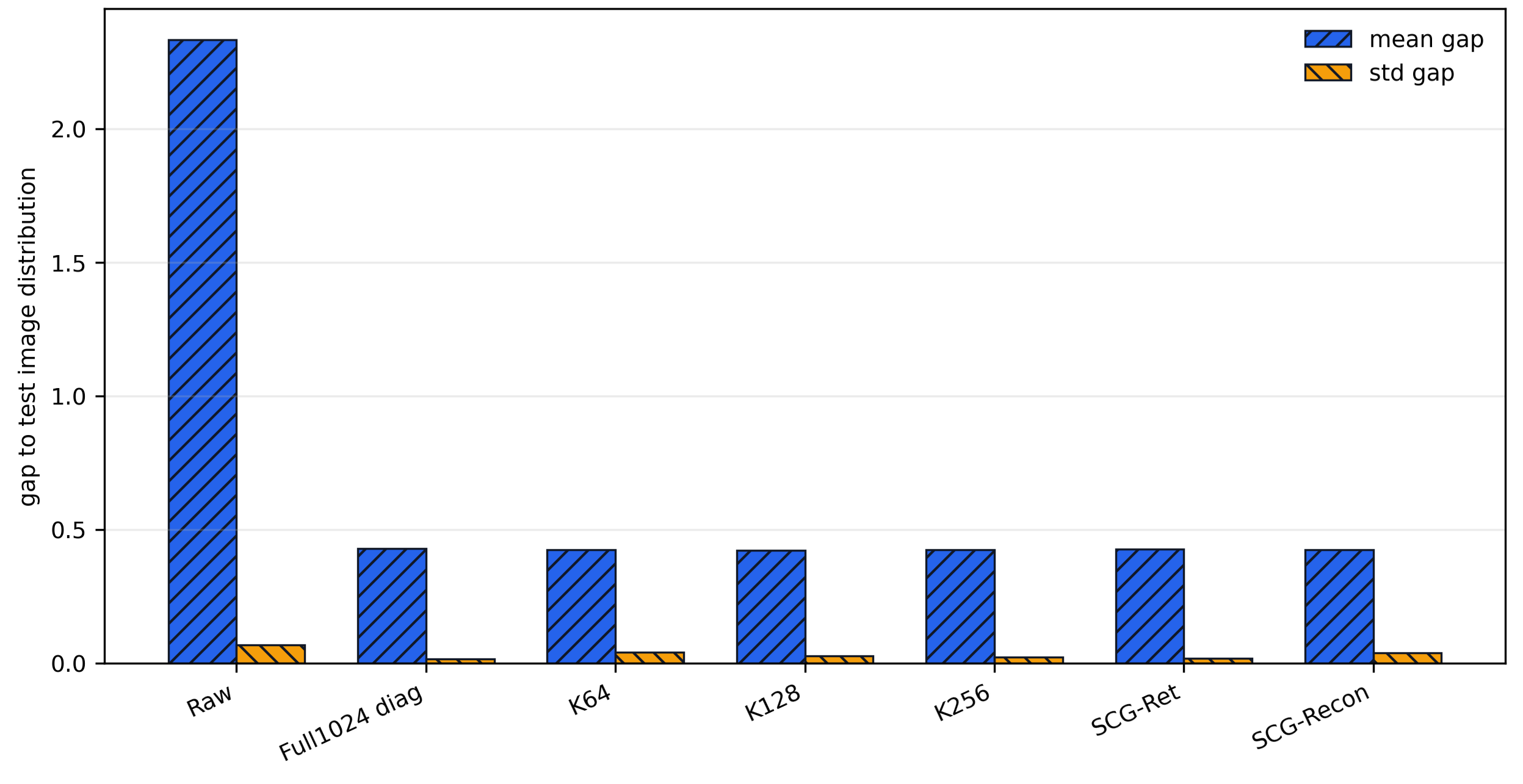}
\caption{Cross-subject Sub-08 distribution alignment.}
\label{fig:distribution-gap}
\end{figure}

\begin{figure}[!t]
\centering
\includegraphics[width=\linewidth]{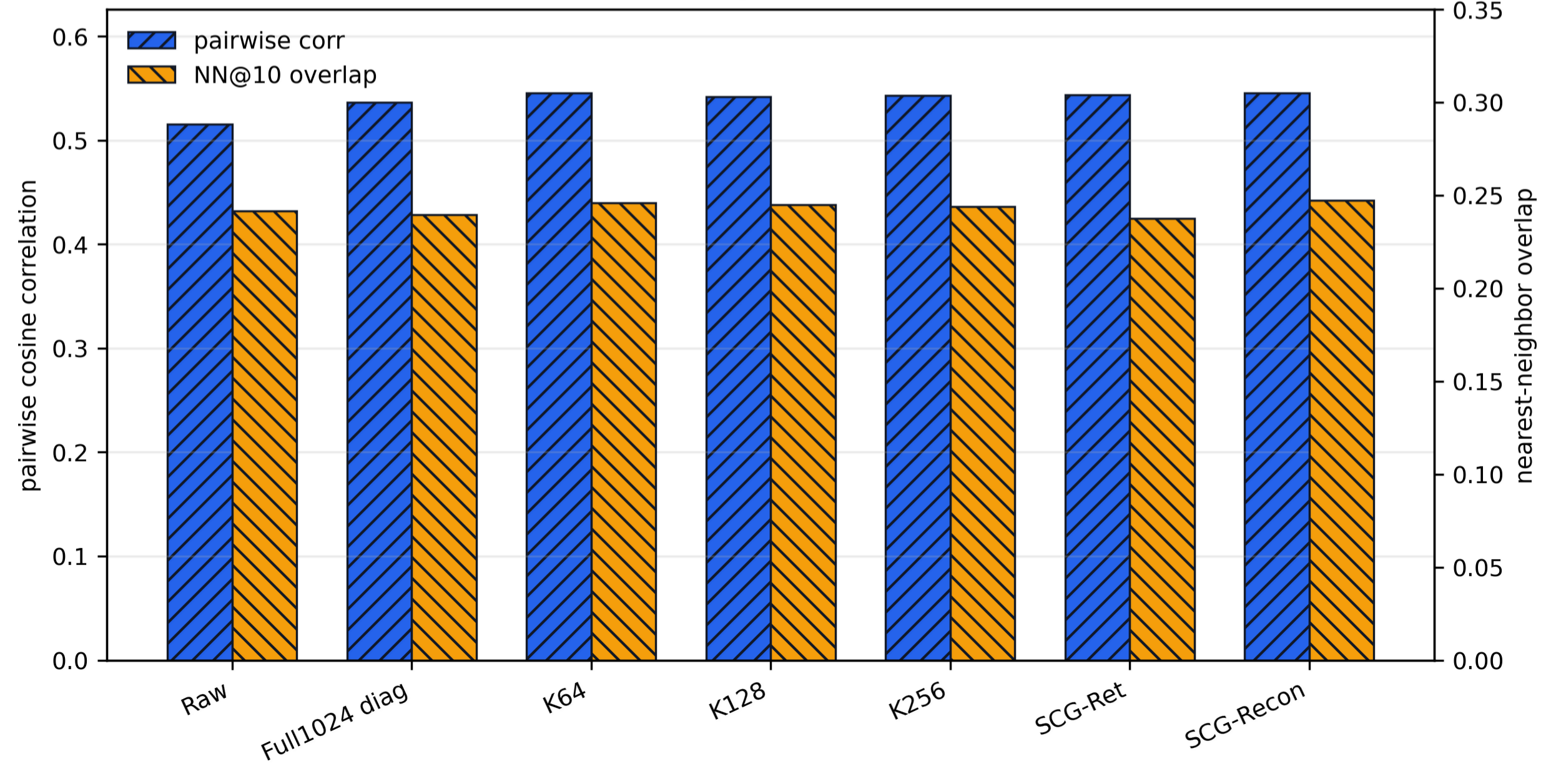}
\caption{Cross-subject Sub-08 geometry preservation.}
\label{fig:geometry-preservation}
\end{figure}

\begin{figure}[!t]
\centering
\includegraphics[width=\linewidth]{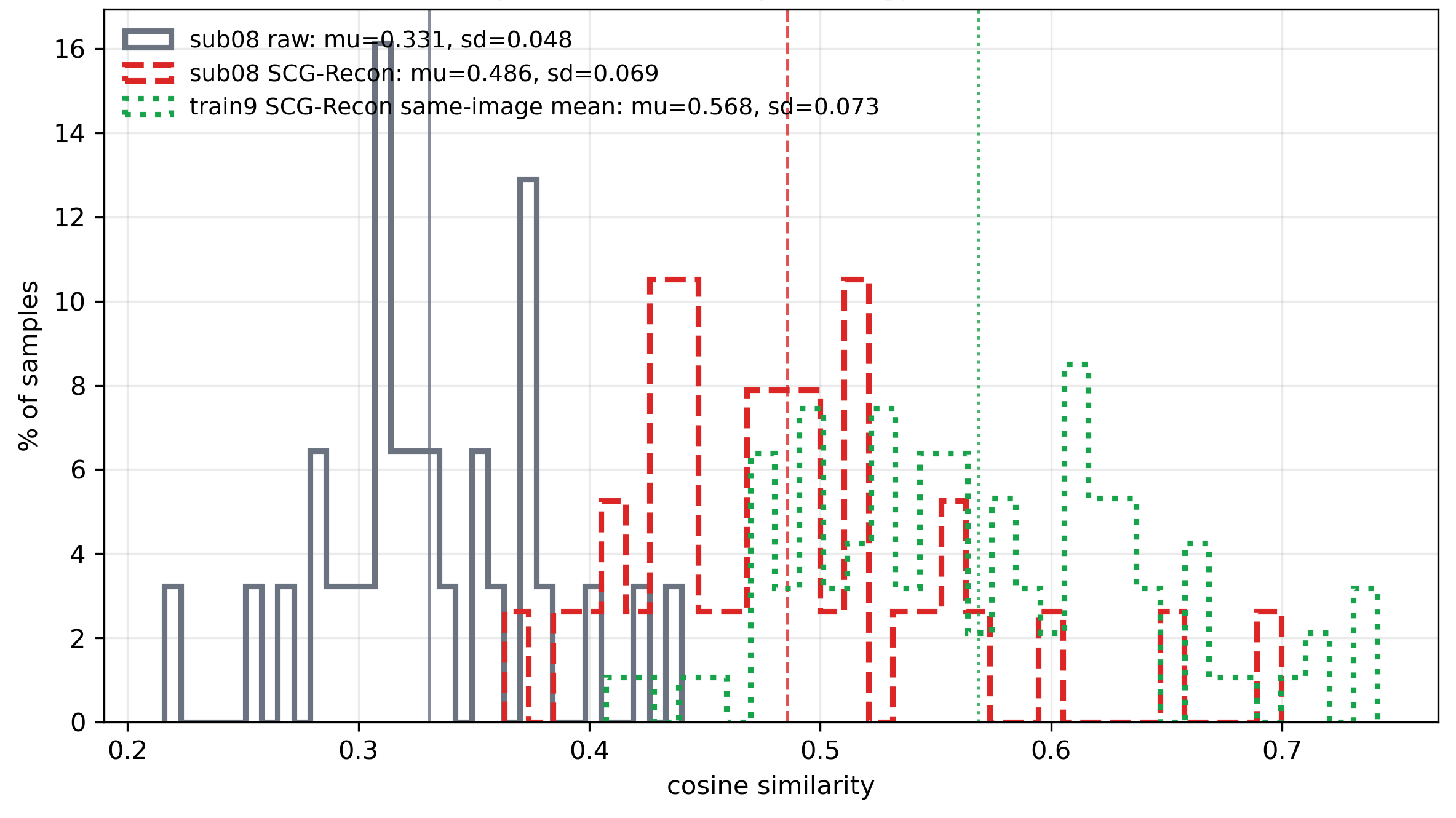}
\caption{Cross-subject Sub-08 alignment to the image side on Top-1 success samples.}
\label{fig:top1-success-cosine}
\end{figure}

Figure~\ref{fig:selector-composition} indicates that the confidence selector routes \retreadout{} primarily to the high-rank K256 candidate, while \reconreadout{} is routed mainly to K64.
Figure~\ref{fig:distribution-gap} indicates that full-dimensional diagonal (Full1024 diag) alignment removes most global mean/std mismatch, while Figure~\ref{fig:geometry-preservation} shows broadly preserved pairwise geometry with limited local nearest-neighbor overlap.
Figure~\ref{fig:top1-success-cosine} repeats the cosine analysis against the paired image on samples with successful Top-1 retrieval.
The oracle rows in Tables~\ref{tab:selector-ablation-recon} quantify the remaining selection gap, whereas the confidence rows correspond to the label-free readouts used at inference.

\subsection{Prior Replacement Diagnostic}

\begin{figure}[!t]
\centering
\includegraphics[width=\linewidth]{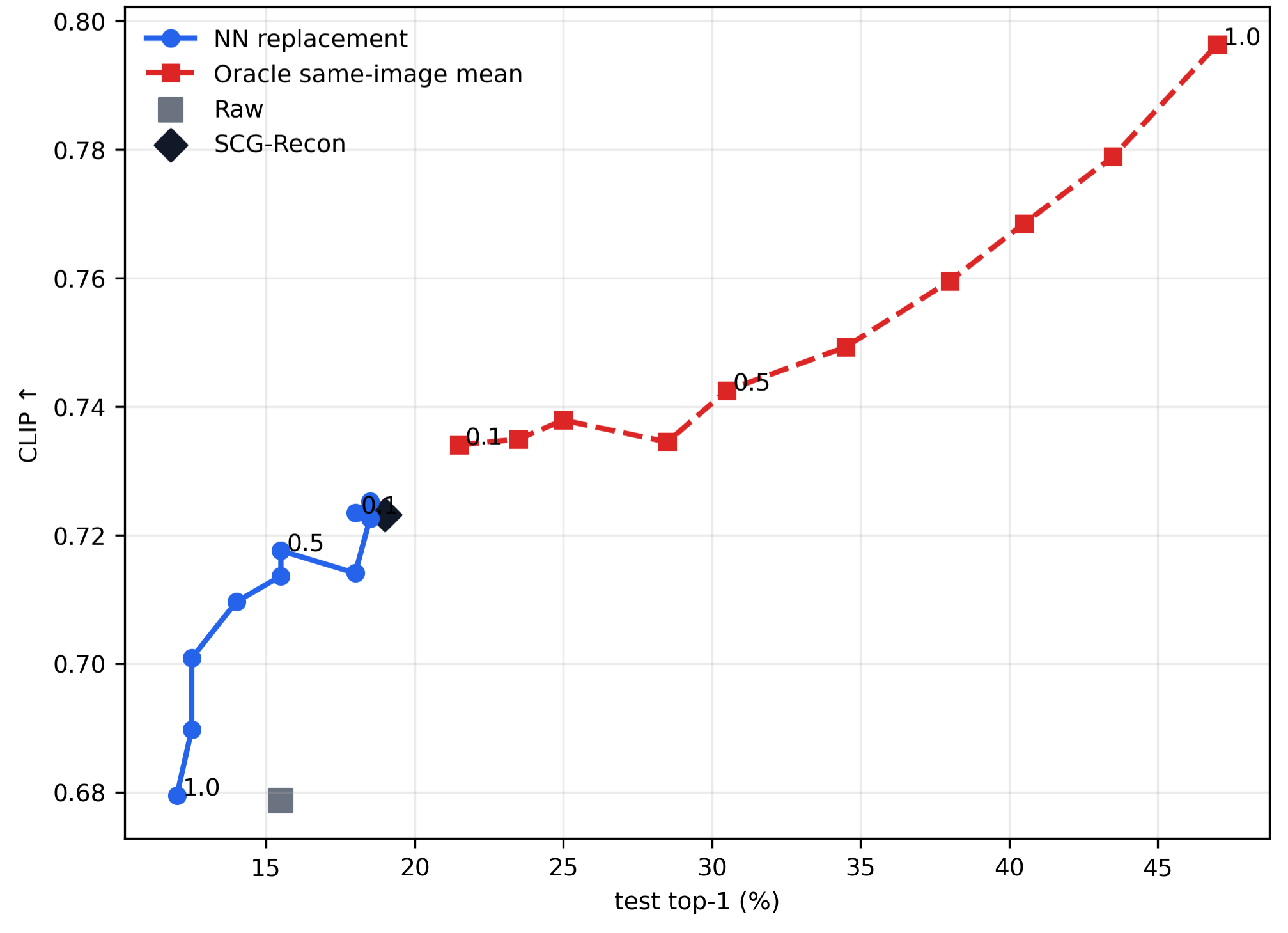}
\caption{Cross-subject Sub-08 prior replacement trade-off.}
\label{fig:prior-replacement-pareto}
\end{figure}

\begin{table}[!t]
\centering
\small
\setlength{\tabcolsep}{1pt}
\begin{tabular}{@{}lcccccc@{}}
\toprule
Method & Ratio & Top-1 $\uparrow$ & CLIP $\uparrow$ & Alex.2 $\uparrow$ & Alex.5 $\uparrow$ & SwAV $\downarrow$ \\
\midrule
Base+\omcr{} & -- & 15.5 & 0.679 & 0.676 & 0.680 & 0.631 \\
\reconreadout{} & -- & 19.0 & 0.723 & 0.684 & 0.726 & 0.618 \\
NN replace & 0.200 & 18.5 & 0.725 & 0.689 & 0.734 & 0.612 \\
NN replace & 1.00 & 12.0 & 0.679 & 0.697 & 0.723 & 0.625 \\
Oracle & 0.500 & \underline{30.5} & \underline{0.743} & \underline{0.712} & \underline{0.756} & \underline{0.595} \\
Oracle & 1.00 & \textbf{47.0} & \textbf{0.796} & \textbf{0.765} & \textbf{0.828} & \textbf{0.564} \\
\bottomrule
\end{tabular}
\caption{Cross-subject Sub-08 prior replacement analysis.}
\label{tab:prior-replacement}
\end{table}

The replacement study probes reusable cross-subject image-specific structure through controlled source-subject substitution.
Table~\ref{tab:prior-replacement} reports the numeric prior-replacement results corresponding to Figure~\ref{fig:prior-replacement-pareto}.
Nearest-neighbor replacement provides a modest reconstruction gain at a moderate replacement ratio, whereas full nearest-neighbor replacement reduces Top-1 retrieval.
Oracle replacement improves all semantic reconstruction metrics by using source-subject responses to the same image, thereby estimating the amount of image-specific structure available across subjects.
Figure~\ref{fig:prior-replacement-pareto} plots the same intervention as retrieval--reconstruction points: blue points are NN replacement ratios, red points are oracle means over responses to the same image, the gray square is Raw, and the black diamond is \reconreadout{}.
The plot places \reconreadout{} near the best NN reconstruction point while retaining higher retrieval than high-ratio NN replacement.
Together, the table and plot indicate that direct source substitution and target-subject calibration capture different cross-subject signals, with the oracle row serving as an analysis reference for shared image-specific information.

\subsection{Implementation Sensitivity Checks}

\begin{table}[!t]
\centering
\small
\setlength{\tabcolsep}{5pt}
\begin{tabular}{lcc}
\toprule
$\beta$ & Top-1 (\%) $\uparrow$ & Top-5 (\%) $\uparrow$ \\
\midrule
0.0625 & \underline{22.5} & 45.5 \\
0.125 & \textbf{24.0} & \underline{46.0} \\
0.250 & \underline{22.5} & \textbf{46.5} \\
0.500 & 20.0 & 45.5 \\
\bottomrule
\end{tabular}
\caption{Branch blend ($\beta$) sensitivity.}
\label{tab:blend-scan}
\end{table}

\begin{table}[!t]
\centering
\small
\setlength{\tabcolsep}{1.2pt}
\begin{tabular}{lcccc}
\toprule
$\alpha$ & CLIP $\uparrow$ & Alex.2 $\uparrow$ & Alex.5 $\uparrow$ & SwAV $\downarrow$ \\
\midrule
\retreadout{}:0.10 & \textbf{0.694} & \underline{0.678} & \underline{0.702} & 0.622 \\
0.00 & 0.690 & 0.676 & 0.699 & \textbf{0.620} \\
0.05 & \underline{0.693} & 0.674 & \textbf{0.708} & \underline{0.621} \\
0.20 & 0.688 & \textbf{0.679} & 0.700 & 0.622 \\
\bottomrule
\end{tabular}
\caption{Image teacher interpolation ($\alpha$) sensitivity.}
\label{tab:imgmix-scan}
\end{table}

Tables~\ref{tab:blend-scan} and~\ref{tab:imgmix-scan} record SCG sensitivity runs within the same Sub-08 setting.
The results show stable behavior around the reported configuration and serve as implementation diagnostics for the candidate construction.

\section{E. Intra-Subject and Cross-Subject Reconstruction Metrics}
\label{sec:complete-reconstruction-metrics}

\begin{table}[!t]
\centering
\small
\setlength{\tabcolsep}{2pt}
\begin{tabular}{lcccccc}
\toprule
Subject & PixCorr $\uparrow$ & SSIM $\uparrow$ & Alex.2 $\uparrow$ & Alex.5 $\uparrow$ & CLIP $\uparrow$ & SwAV $\downarrow$ \\
\midrule
Sub-01 & 0.152 & 0.350 & 0.678 & 0.729 & 0.703 & 0.628 \\
Sub-02 & 0.149 & 0.367 & \textbf{0.752} & \underline{0.763} & 0.710 & \underline{0.607} \\
Sub-03 & 0.128 & 0.374 & 0.730 & 0.750 & \underline{0.732} & 0.614 \\
Sub-04 & 0.138 & 0.348 & 0.670 & 0.719 & 0.693 & 0.630 \\
Sub-05 & \underline{0.159} & 0.366 & 0.720 & 0.753 & 0.696 & 0.623 \\
Sub-06 & 0.154 & 0.368 & 0.678 & 0.730 & 0.704 & 0.612 \\
Sub-07 & 0.134 & 0.347 & 0.682 & 0.736 & 0.703 & 0.632 \\
Sub-08 & \textbf{0.174} & \textbf{0.389} & 0.684 & 0.726 & 0.723 & 0.618 \\
Sub-09 & 0.156 & \underline{0.381} & 0.723 & 0.744 & 0.682 & 0.618 \\
Sub-10 & 0.147 & 0.359 & \underline{0.737} & \textbf{0.803} & \textbf{0.781} & \textbf{0.601} \\
\midrule
Mean & 0.149 & 0.365 & 0.706 & 0.745 & 0.713 & 0.618 \\
\bottomrule
\end{tabular}
\caption{Cross-subject \reconreadout{} under ten-subject LOSO.}
\label{tab:all-subjects-inter-scgrecon}
\end{table}

\begin{table}[!t]
\centering
\small
\setlength{\tabcolsep}{1mm}
\begin{tabular}{@{}llcccc@{}}
\toprule
Method & Stat & Alex.2 $\uparrow$ & Alex.5 $\uparrow$ & CLIP $\uparrow$ & SwAV $\downarrow$ \\
\midrule
Base+\omcr{} & Mean & 0.681 & 0.701 & 0.656 & 0.640 \\
 & Std & 0.012 & 0.023 & 0.030 & 0.016 \\
\fulldiag{} & Mean & \underline{0.699} & \underline{0.731} & \underline{0.695} & \underline{0.623} \\
 & Std & 0.023 & 0.024 & 0.035 & 0.014 \\
\retreadout{} & Mean & 0.695 & 0.724 & 0.688 & 0.625 \\
 & Std & 0.023 & 0.031 & 0.038 & 0.012 \\
\reconreadout{} & Mean & \textbf{0.706} & \textbf{0.745} & \textbf{0.713} & \textbf{0.618} \\
 & Std & 0.028 & 0.023 & 0.027 & 0.010 \\
\bottomrule
\end{tabular}
\caption{Ten-subject LOSO cross-subject reconstruction summary, reported as mean and population standard deviation.}
\label{tab:inter-recon}
\end{table}

\begin{table}[!t]
\centering
\small
\setlength{\tabcolsep}{1.5pt}
\begin{tabular}{lcccccc}
\toprule
Subject & PixCorr $\uparrow$ & SSIM $\uparrow$ & Alex.2 $\uparrow$ & Alex.5 $\uparrow$ & CLIP $\uparrow$ & SwAV $\downarrow$ \\
\midrule
Sub-01 & 0.252 & 0.382 & 0.849 & 0.879 & 0.812 & 0.583 \\
Sub-02 & 0.242 & 0.384 & 0.829 & 0.884 & 0.835 & 0.564 \\
Sub-03 & 0.249 & 0.375 & 0.845 & 0.872 & 0.858 & 0.570 \\
Sub-04 & 0.228 & 0.371 & 0.832 & 0.880 & 0.861 & 0.556 \\
Sub-05 & 0.223 & 0.368 & 0.834 & 0.858 & 0.809 & 0.579 \\
Sub-06 & 0.261 & 0.376 & \underline{0.870} & 0.901 & 0.858 & 0.555 \\
Sub-07 & 0.230 & 0.372 & 0.819 & 0.862 & 0.859 & 0.561 \\
Sub-08 & \textbf{0.304} & \textbf{0.397} & 0.868 & \underline{0.905} & \textbf{0.888} & \underline{0.538} \\
Sub-09 & \underline{0.264} & \underline{0.387} & 0.836 & 0.875 & 0.827 & 0.560 \\
Sub-10 & 0.262 & 0.382 & \textbf{0.871} & \textbf{0.920} & \underline{0.886} & \textbf{0.531} \\
\midrule
Mean & 0.251 & 0.380 & 0.845 & 0.883 & 0.849 & 0.560 \\
\bottomrule
\end{tabular}
\caption{Intra-subject \reconreadout{}.}
\label{tab:all-subjects-scgrecon}
\end{table}

These tables provide the complete reconstruction results associated with the main decoding analysis.
Tables~\ref{tab:all-subjects-inter-scgrecon} and~\ref{tab:inter-recon} report cross-subject LOSO reconstruction by subject and by SCG readout, while Table~\ref{tab:all-subjects-scgrecon} reports the corresponding intra-subject \reconreadout{} results.

\section{F. Intra-Subject and Cross-Subject Retrieval Accuracy}
\label{sec:appendix-retrieval}

\paragraph{Compact retrieval summary.}
Retrieval provides an additional measure of semantic alignment.
Table~\ref{tab:retrieval} reports the ten-subject mean and population standard deviation, and Table~\ref{tab:retrieval-full-intra} provides the intra-subject breakdown by subject.
NeuroBridge (ViT-H-14) is evaluated in the same semantic space used by our reconstruction pipeline.

\begin{table}[!t]
\centering
\small
\setlength{\tabcolsep}{1.5pt}
\begin{tabular}{@{}lcccc@{}}
\toprule
Method & \multicolumn{2}{c}{Intra-subject} & \multicolumn{2}{c}{Cross-subject} \\
\cmidrule(lr){2-3}\cmidrule(lr){4-5}
 & Top-1 (\%) $\uparrow$ & Top-5 (\%) $\uparrow$ & Top-1 (\%) $\uparrow$ & Top-5 (\%) $\uparrow$ \\
\midrule
NICE & 16.1 $\pm$ 3.51 & 43.6 $\pm$ 6.52 & 6.20 $\pm$ 1.06 & 21.4 $\pm$ 2.82 \\
ATM & 27.1 $\pm$ 7.02 & 58.1 $\pm$ 7.74 & 11.9 $\pm$ 4.56 & 33.8 $\pm$ 8.27 \\
AVDE & 30.0 $\pm$ 5.90 & 58.2 $\pm$ 7.60 & 14.3 $\pm$ 3.57 & 32.9 $\pm$ 6.61 \\
NeuroBridge \\ (ViT-H-14) & 50.9 $\pm$ 6.14 & 82.4 $\pm$ 6.73 & 16.4 $\pm$ 4.20 & 39.5 $\pm$ 6.64 \\
\midrule
\fulldiag{} & 51.2 $\pm$ 6.24 & \textbf{83.9} $\pm$ 5.55 & 22.5 $\pm$ 3.88 & \textbf{50.1} $\pm$ 5.72 \\
\retreadout{} & \textbf{52.8} $\pm$ 6.49 & \underline{83.6} $\pm$ 5.91 & \textbf{23.7} $\pm$ 3.80 & 49.2 $\pm$ 5.72 \\
\reconreadout{} & \underline{51.9} $\pm$ 5.97 & 82.8 $\pm$ 6.04 & \underline{23.3} $\pm$ 3.87 & \underline{49.5} $\pm$ 4.46 \\
\bottomrule
\end{tabular}%
\caption{Mean intra-subject and cross-subject retrieval accuracy, reported as mean $\pm$ population standard deviation.}
\label{tab:retrieval}
\end{table}

In the cross-subject setting, \retreadout{} achieves the highest mean Top-1 accuracy, while the diagonal candidate gives the highest Top-5 accuracy.

\begin{table*}[!t]
\centering
\small
\setlength{\tabcolsep}{1mm}
\begin{tabular*}{\textwidth}{@{\extracolsep{\fill}}ll|ccccccccccc@{}}
\toprule
Method & Metric (\%) & S01 & S02 & S03 & S04 & S05 & S06 & S07 & S08 & S09 & S10 & Mean \\
\midrule
\multicolumn{13}{c}{Intra-subject: train and test on one subject} \\
\midrule
NICE & Top-1 & 13.2 & 13.5 & 14.5 & 20.6 & 10.1 & 16.5 & 17.0 & 22.9 & 15.4 & 17.4 & 16.1 \\
 & Top-5 & 39.5 & 40.3 & 42.7 & 52.7 & 31.5 & 44.0 & 42.1 & 56.1 & 41.6 & 45.8 & 43.6 \\
\midrule
ATM & Top-1 & 25.6 & 22.0 & 25.0 & 31.4 & 12.9 & 21.3 & 30.5 & 38.8 & 34.4 & 29.1 & 27.1 \\
 & Top-5 & 60.4 & 54.5 & 62.4 & 60.9 & 43.0 & 51.1 & 61.5 & 72.0 & 51.5 & 63.5 & 58.1 \\
\midrule
AVDE & Top-1 & 25.0 & 24.1 & 27.5 & 29.8 & 25.4 & 33.5 & 27.4 & 41.7 & 26.1 & 39.5 & 30.0 \\
 & Top-5 & 55.2 & 51.0 & 58.6 & 54.7 & 50.3 & 60.3 & 55.2 & 71.3 & 52.1 & 73.0 & 58.2 \\
\midrule
NeuroBridge (ViT-H-14) & Top-1 & \underline{43.5} & \textbf{53.5} & \underline{51.0} & \underline{52.0} & \underline{41.0} & \underline{56.0} & \underline{47.5} & \underline{58.5} & \underline{45.5} & \textbf{60.5} & \underline{50.9} \\
 & Top-5 & \underline{71.5} & \underline{79.0} & \textbf{86.0} & \textbf{83.5} & \underline{70.5} & \underline{87.5} & \underline{83.5} & \underline{88.5} & \textbf{81.5} & \textbf{92.5} & \underline{82.4} \\
\midrule
\retreadout{} (Ours) & Top-1 & \textbf{48.5} & \underline{52.0} & \textbf{52.5} & \textbf{53.5} & \textbf{41.5} & \textbf{60.5} & \textbf{49.0} & \textbf{63.5} & \textbf{47.0} & \underline{60.0} & \textbf{52.8} \\
 & Top-5 & \textbf{77.5} & \textbf{81.0} & \underline{85.5} & \textbf{83.5} & \textbf{71.0} & \textbf{89.5} & \textbf{86.5} & \textbf{89.5} & \underline{80.5} & \underline{91.0} & \textbf{83.6} \\
\bottomrule
\end{tabular*}
\caption{Complete intra-subject 200-way retrieval accuracy.}
\label{tab:retrieval-full-intra}
\end{table*}

\section{G. Additional Qualitative Reconstructions}
\begin{figure}[!t]
\centering
\includegraphics[width=\linewidth,height=0.70\textheight,keepaspectratio]{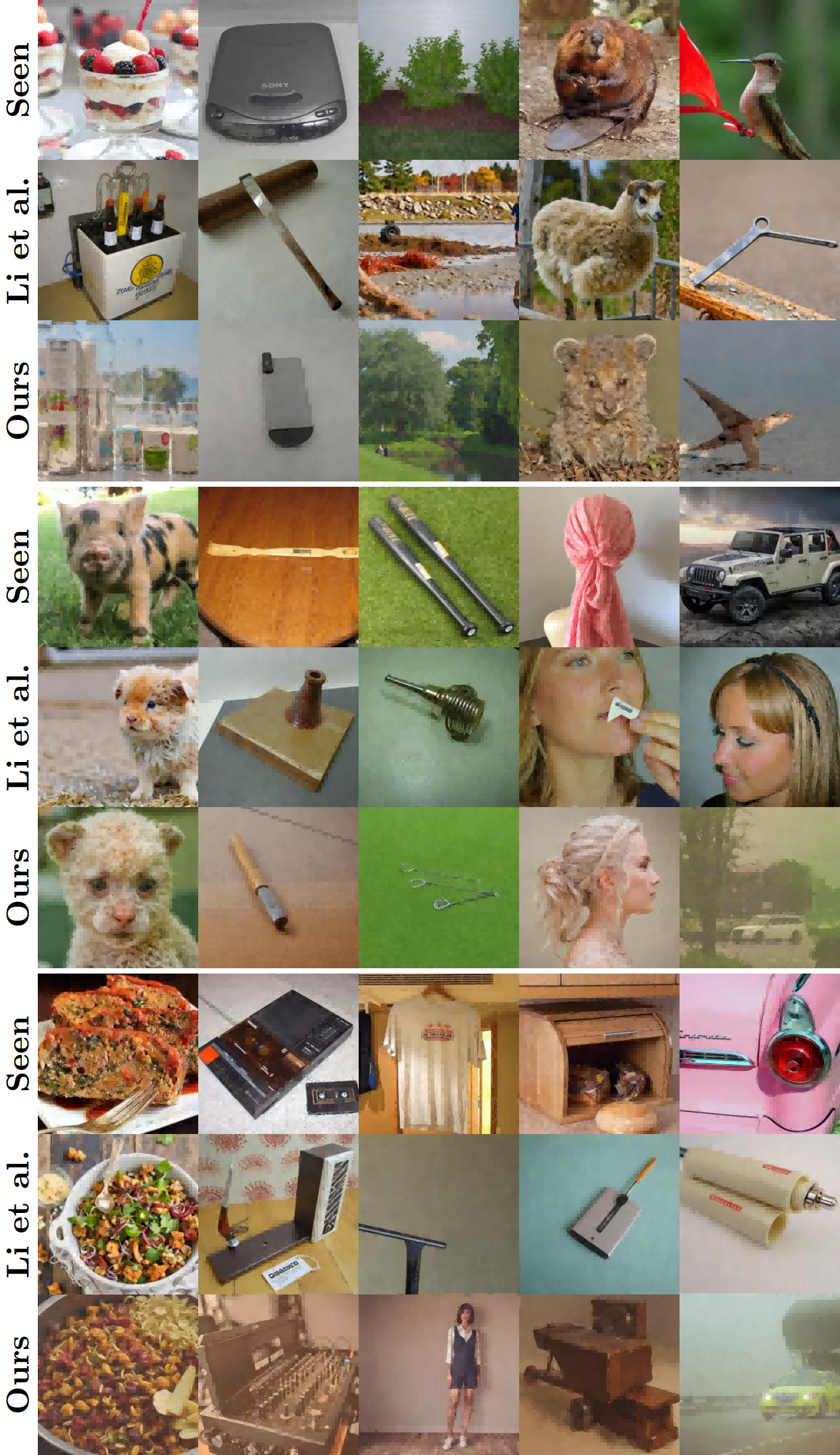}
\caption{Additional cross-subject Sub-08 qualitative reconstruction examples.}
\label{fig:inter-qual-appendix}
\end{figure}

\begin{figure}[!t]
\centering
\includegraphics[width=\linewidth]{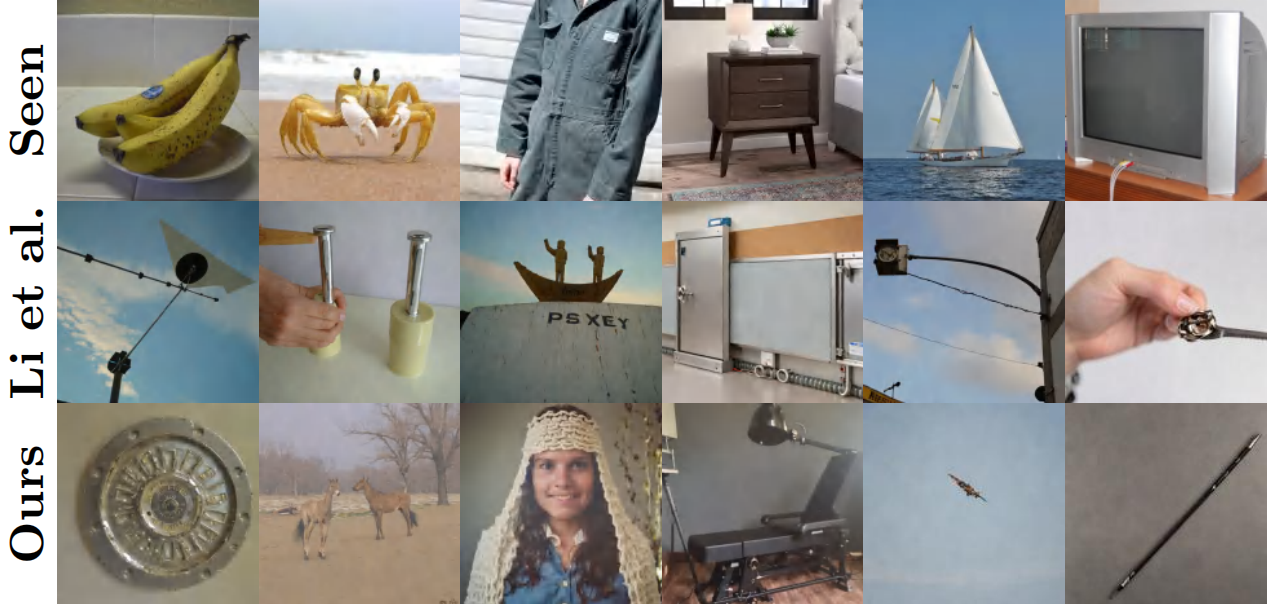}
\caption{Representative difficult cases from the remaining cross-subject Sub-08 reconstructions. Rows compare the seen stimulus, Li et al., and our reconstruction.}
\label{fig:inter-badcases-appendix}
\end{figure}

\paragraph{Limitations and future work.}
Figure~\ref{fig:inter-qual-appendix} extends the qualitative comparison across food, plants, animals, objects, clothing, and transportation, whereas Figure~\ref{fig:inter-badcases-appendix} presents challenging samples selected from the remaining reconstructions.
Errors range from complete category loss to preservation of a coarse cue, such as clothing or transportation, without the object identity.
Furniture and electronic devices further expose weak recovery of target-specific appearance and geometry.
The visual coherence of these outputs, despite their semantic mismatch, shows that image realism alone does not ensure semantic fidelity when fine-grained EEG evidence from a held-out subject is ambiguous.

Future work may refine both the neural representation and its interface with generation.
Modeling EEG frequency and temporal dynamics with electrode topology, together with reliability-weighted trial aggregation, could stabilize the inputs to \scgabbr{}.
Candidate weighting informed by uncertainty and multiscale coordinates could improve semantic precision, while spatial guidance in the low-level branch could preserve object geometry.
Additional subjects and recording sessions would assess these extensions under zero-shot transfer.

\end{document}